\newcount\formattingstyle
\newcommand{\docBeginsHeader}{}
\newcommand{\putReferencesList}{
    \bibliography{references}
    \bibliographystyle{unsrt}            
}
\newcommand{\EqLineBreak}{ \\ }

\ifcase\formattingstyle
    \newcommand{\putDocHeader}{
        \documentclass[final,3p,times,twocolumn]{elsarticle}

        \usepackage{amsthm}
        \newtheorem{remark}{Remark}
        \newtheorem{proposition}{Proposition}
        \usepackage{algorithm}
        \usepackage{algpseudocode}        
    }
    \newcommand{\orgname}[1]{#1}    
    \newcommand{\orgaddress}[1]{#1} 
    
    \newcommand{\abstractText}[1]{
        \begin{abstract}
        #1
        \end{abstract}
        
        \maketitle
    }
    
\or 
      \newcommand{\putDocHeader}{
        \documentclass{article}
        \usepackage{authblk}
      }
    \newcommand{\corref}{}
    \newcommand{\cortext}{}
    \newcommand{\address}{\affil}
    \newcommand{\ead}{}
\or 
    \newcommand{\putDocHeader}{
        \documentclass[AMA,Times2COL]{WileyNJDv5}

    }
    \newcommand{\corref}[1]{}
    \newcommand{\cortext}[2][default]{ 
    }
    \newcommand{\ead}[1]{}

    \renewcommand{\putReferencesList}{
        \bibliography{references}    
    }
    \newcommand{\abstractText}[1]{
        \abstract{#1}
        
        \maketitle
    }

    \renewcommand{\docBeginsHeader}{

        \articletype{} 
        
        \journal{}
        \volume{}
        \copyyear{}
        \startpage{1}            
        
    }

\or 

    \newcommand{\putDocHeader}{
        \documentclass[12pt]{article} 
        \usepackage{setspace}         
        \usepackage{geometry}         

        \usepackage{float}

        \geometry{
            top=1in,
            bottom=1in,
            left=1in,
            right=1in
        }        
        \usepackage{url}
        \usepackage{amsthm}
        
        \newtheorem{proposition}{Proposition}
        \usepackage{algorithm}
        \usepackage{algpseudocode}        

        \usepackage{authblk}

        \renewcommand{\Affilfont}{\small}
    
    }
    \renewcommand{\docBeginsHeader}{
        \doublespacing 
    }

    \newcommand{\corref}[1]{\textsuperscript{}}
    \newcommand{\cortext}[2][default]{ 
    }
    \newcommand{\abstractText}[1]{
        \maketitle
    
        \begin{abstract}
        #1
        \end{abstract}
    }

    \newcommand{\address}{\affil}
    
    \newcommand{\ead}[1]{}

    \newcommand{\orgname}[1]{#1}    
    \newcommand{\orgaddress}[1]{#1} 

    \renewcommand{\EqLineBreak}{ }

\fi

\makeatletter
\newcommand{\halfpagefgraphics}[1][]{
    \if@twocolumn
        \includegraphics[width=0.99\linewidth]{#1}
    \else
    {
        \includegraphics[width=0.5\linewidth]{#1}
    }
    \fi
}
\makeatother

\putDocHeader

\usepackage[english]{babel}
\usepackage[utf8]{inputenc}
\usepackage{amsmath,amssymb,graphicx,xcolor,tabularx}

\usepackage{cuted} 
\usepackage{flafter}

\usepackage{multirow}

\newcommand{\RemoveForNow}[1]{}

\newcommand{\assign}{:=}
\newcommand{\cdummy}{\cdot}
\newcommand{\textdots}{...}
\newcommand{\tmcolor}[2]{{\color{#1}{#2}}}
\newcommand{\tmop}[1]{\ensuremath{\operatorname{#1}}}
\newcommand{\tmrsub}[1]{\ensuremath{_{\textrm{#1}}}}
\newcommand{\tmrsup}[1]{\textsuperscript{#1}}
\newcommand{\tmtextit}[1]{\text{{\itshape{#1}}}}

\begin{document}

\docBeginsHeader

\title{Physics-informed neural networks need a physicist to be accurate: the case of mass and heat transport in Fischer-Tropsch catalyst particles}

\author[1,2]{Tymofii Nikolaienko\corref{cor1}}
\ead{tniko@softserveinc.com}

\author[3]{Harshil Patel\corref{cor1}}

\author[3]{Aniruddha Panda\corref{cor1}}
\ead{Aniruddha.Panda@shell.com}

\author[3]{Subodh Madhav Joshi}

\author[4]{Stanislav Jaso}

\author[3]{Kaushic Kalyanaraman}

\cortext[cor1]{Corresponding author}

\address[1]{\orgname{SoftServe Inc.}, \orgaddress{2d Sadova St., 79021 Lviv, Ukraine}}
\address[2]{\orgname{Taras Shevchenko National University of Kyiv}, \orgaddress{64/13 Volodymyrska Str., Kyiv 01601, Ukraine}}
%


\address[3]{\orgname{Shell India Markets Pvt. Ltd. (Shell Projects \& Technology)}, \orgaddress{Mahadeva Kodigehalli, Bengaluru, Karnataka 562149, India}}
\address[4]{\orgname{Shell Global Solutions International B.V.}, \orgaddress{Grasweg 31, 1031 HW Amsterdam, The Netherlands}}

\abstractText{
  Physics-Informed Neural Networks (PINNs) have emerged as an influential technology, merging the swift and automated capabilities of machine learning with the precision and dependability of simulations grounded in theoretical physics.
  PINNs are often employed to solve algebraic or differential equations  to replace some or even all steps of multi-stage computational  workflows, leading to their significant speed-up.
  However, wide adoption of PINNs is still hindered by reliability issues, particularly at extreme ends of the input parameter ranges. 
  In this study, we demonstrate this in the context of a system of coupled  non-linear differential reaction-diffusion and heat transfer equations related to Fischer-Tropsch  synthesis, which are solved by a finite-difference method with a PINN used in evaluating their source terms.
  It is shown that the testing strategies traditionally used to assess the  accuracy of neural networks as function approximators can overlook the  peculiarities which ultimately cause instabilities of the finite-difference  solver.  
  We propose a domain knowledge-based modifications to the PINN architecture  ensuring its correct asymptotic behavior.  
  When combined with an improved numerical scheme employed as an initial guess  generator, the proposed modifications are shown to recover the overall  stability of the simulations, while preserving the speed-up brought by PINN as the workflow component.  
  We discuss the possible applications of the proposed hybrid  transport equation solver in context of chemical reactors simulations.
}



\section{Introduction}

The outstanding abilities of neural networks (NNs) in approximating complex relations have resulted in their successful application in many fields, ranging from image recognition and text comprehension to mimicking the solutions of differential equations encountered in complex engineering problems \cite{Cuomo2022}.
One of the benefits brought by employing NNs as an alternative to traditional numerical methods is shifting the computational burden to the training phase, which is performed only once, thus enabling faster solution generation during the inference phase. 
This can be especially helpful in accelerating multi-stage
simulations when the output of one computational method is used as an input to another one, as often encountered in engineering problems or digital twins designs \cite{Tao2019,herwig2022}.

An illustrative example can be found in chemical engineering problems related to ground-up modeling of chemical reactor or even entire chemical plants.
In such applications, theoretical models are commonly available for finding the rates of both the micro-scale phenomena (e.g., molecular-level chemical reactions) and macro-scale phenomena (e.g., heat and mass transport). 
Their coupling results then in a system of equations which should be solved self-consistently, e.g., by solving the `micro-scale' equations as a sub-task each time when the evaluation of the source terms in `macro-scale' equations is required.
Replacing solution of such sub-tasks with NNs is then an attractive option to accelerate the overal simulation.

Despite their advantages, NNs, like many other models which learn from data, often lack interpretability.
This makes their reliability in scientific or mission-critical applications questionable. 
Physics-informed neural network (PINN) approach has been proposed to partially overcome this drawback by incorporating the available theoretical knowledge into the NN training process \cite{Raissi2019}. This approach suggests using exact equations known from the theory as the objectives that the function approximated by NN is expected to satisfy.
Imposing such type of constraints often appears sufficient to make NN fit the solution of a theory-based (typically, physics-based) equation and in this way to become more interpretable.
The physics-informed paradigm also requires minimal changes to the NN architecture and can be conveniently implemented in one of numerous specialized programming frameworks.
However, as the loss function which is minimized during any NN training does not commonly reach exactly zero during or after this process, the theory-based constraints can only be satisfied approximately.
Thus, the transfer of the theoretical knowledge into the PINN achieved by the physics-informed method remains incomplete.
As will be discussed below, this can have significant consequences for incorporating PINNs into the multi-stage simulations.

More broadly, PINNs can be viewed in context of wider family of methods,
commonly known as the physics-informed machine learning (PIML).
Within this family, there are other methods which are also intended to fuse theory-based and data-driven methods into a single computational model, but achieve this by modifying the architecture of the NN itself in order to make its output by design fulfil certain theory-based constraints exactly (e.g., \cite{NEURIPS2018_a3fc981a, TFN-2018-arXiv, SE3-transf-2020, pmlr-v139-satorras21a}).
Although none of the PIML approaches typically achieves complete transfer of all available theoretical knowledge into the NN, different approaches prioritize different aspects of it.
To make this point more concrete, it is instructive to distinguish between the accuracy of the numerical values produced by NNs and the correctness of their dependence on the NN input parameters on the asymptotics.

In this paper, we demonstrate that minor numerical inaccuracies of PINN as a function approximator can significantly affect the overall result of a multi-stage simulation, when the PINN acts as a source term in a diffusion-like equation (subsection \ref{model-example-subsection}).
We further investigate the asymptotics suggested by the theory-based equations (subsection \ref{subsection-asympt-an}) and propose a modified PINN architecture which ensures the model follows them by design (subsections \ref{subsection-SPINN-post-processing}, \ref{train-val-subsection}).
This is done for the particular case of reaction-diffusion system related to Fischer-Tropsch synthesis (FTS) process, which is widely used in chemical industry to produce synthetic hydrocarbons (the underlying equations are reviewed in subsections \ref{section-microkinetics-equations} and \ref{subsection-eqs-intro}, along with the benefits of leveraging PINN for solving them).
Finally, we demonstrate that a well-defined and guaranteed asymptotic behavior of a modified PINN is essential for constructing a conventional finite-difference equation solver enabling a stable convergence for the considered problem (subsection \ref{appendix-conv-solver-initial-guess}).

\section{Defining the problem} \label{sect-theor-and-prelim}

\subsection{
The challenge of simulating the Fisher-Tropsch synthesis process
} \label{section-microkinetics-equations}



As a chemical process used for converting synthesis gas which is a mixture of carbon monoxide (CO) and hydrogen ($\mathrm{H}_2$) into liquid hydrocarbons, Fischer-Tropsch Synthesis (FTS) plays an important role in the production of synthetic fuels and chemicals \cite{van-der-Laan-kinetics-1999, Mendez-kinetic-2020, Martinelli2020, Centi2020}.
It has become an invaluable tool in the chemical industry, offering a pathway to sustainable and diversified fuel sources, providing an alternative route to conventional petrochemical processes.

At its core, the FTS process involves the catalytic reaction, which starts from CO and $\mathrm{H_2}$, and forms a wide range of hydrocarbons, such as alkanes, alkenes.
The overall chemical reactions in which FTS transformations can be summarized are:
\begin{equation}
  n \text{CO} + (2 n + 1)  \text{H}_2 \rightarrow \text{C}_n \text{H}_{2
  \text{n} + 2} + n \text{H}_2 \text{O}  
  \label{gross-paraffin-reaction-eq}
\end{equation}
\begin{equation}
  n \text{CO} + 2 n \hspace{0.27em} \text{H}_2 \rightarrow \text{C}_n
  \text{H}_{2 n} + n \text{H}_2 \text{O}  \label{gross-olefin-reaction-eq}
\end{equation}
Detailed mechanism behind these overall reactions depends on thermodynamic conditions and the type of catalyst, and different models are available to describe their actual chemistry \cite{van-der-Laan-kinetics-1999}.

Catalysts play a central role in the FTS process as the reactions predominantly take place on the active sites which are present on the catalyst surface.
A significant challenge in FTS is optimizing the reaction rates, which is facilitated by maximizing the surface area of these catalysts.
For this purpose, porous catalyst particles are employed, offering a large surface area to volume ratio.
This implies that the simulation of the FTS in practical setups should take both the reactions and species transport into account, thereby forming a reaction-diffusion type of the problem.


Industrial viability of the FTS process is tightly connected with the design of chemical reactors, which provide a controlled environment for the reactions, enhance catalyst performance and manage heat, allowing for efficient, scalable, safe, and flexible production of the hydrocarbons.
Simulating the reactors used for running the FTS process serves multiple important purposes, such as optimizing reaction conditions to maximize their yield and selectivity,
preventing undesirable phenomena, such as thermal runaway, ensuring safety and scalability.


In traditional approaches to simulating FTS reactors, the behavior of catalyst particles is often fitted using empirical formulae. 
However, these models necessitate re-parameterization whenever there are changes in particle characteristics such as, size, shape, or porosity.
Moreover, empirical formulae typically provide reasonable accuracy within the limited range of reaction conditions, which further restricts the flexibility of process optimization under varied operating conditions.

\begin{figure}[b]
\begin{center}
  \raisebox{0.0\height}
{
\halfpagefgraphics[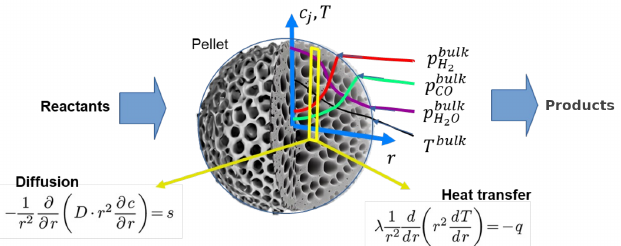]
}
  \caption{
  Schematic representation of the catalytic particle as an element
  of chemical reactor performing Fischer-Tropsch synthesis. The mass and heat
  transport in the particle is modeled by diffusion and heat transfer
  equations, in which the source terms result from the kinetics of underlying
  reactions.
  \label{fig:react-diffus-schematics}
  }
\end{center}
\end{figure}

Alternatively, a ground-up modeling approach allows for a more comprehensive simulation of chemical kinetics at both micro and macro scales.
This approach begins with simulating the fundamental reaction mechanisms and their kinetics at a micro-level, then incorporates these kinetic models into simulations of individual catalyst particles (Fig. \ref{fig:react-diffus-schematics}) and then ends up with the full reactor system.
At all stages, the fundamental relationships known from theory are preferred over empirical ones.
This approach thus provides a more detailed and flexible framework for modeling and optimizing the FTS process.
Moreover, it has the potential for stronger predictive power across a broad range of operating conditions, as it is built on fundamental physical and chemical principles. 
We review the mathematical formulation of the equations underlying the ground-up description of the catalyst particle in context of FTS in the next subsection.
%
%
%

The multi-stage nature of the ground-up approach introduces its own challenges in terms of computational time and numerical stability.
%
%
To tackle some of the these bottlenecks, we've previously leveraged scientific machine learning methods to eliminate the need of solving the microkinetics equations \cite{our-prev-arxiv-preprint}.
We introduced a physics-informed neural network (PINN), which computes the fraction of vacant catalytic sites, a key quantity in FTS mikrokinetics, based on reaction conditions, with median relative error (MRE) of 0.03\%, and the
FTS product formation rates with MRE of 0.1 \%. 
In contrast to conventional equation solvers, PINN model allows natural execution on GPUs, allowing for speedups up to $10^5$.

While the PINN model considerably speeds up microkinetics simulations, its integration into the multi-scale reactor model requires a careful treatment.
As we demonstrate in this study, apart from improving computational efficiency, approximate solutions of the reactions kinetics model equations might still not be directly applicable within a source term in a diffusion equation due to the convergence issues.
We address them in detail in subsections
\ref{model-example-subsection}, 
\ref{subsection-asympt-an},
\ref{subsection-SPINN-post-processing}.
Thus, there exists a need for a mechanism for correct integration of the microkinetics PINN model into the more general multi-scale simulation framework. This forms the primary motivation for this study.  

\subsection{Reaction-diffusion equations in ground-up modelling of
Fisher-Tropsch synthesis process}
\label{subsection-eqs-intro}

The ground-up approach suggests breaking down the overall transformation eqs. \eqref{gross-paraffin-reaction-eq}, \eqref{gross-olefin-reaction-eq} into a series of interconnected elementary steps or stages. 
Each step is then described as an individual reaction with its own rate constant, using which the full reaction pathway kinetics is deduced.
In the case of FTS, the processes described by the elementary steps typically include reactant adsorption, chain initiation, chain growth, and product formation/desorption.
Such a description is typical for a `microkinetic' approach \cite{Notagamwala-microkinetic-2021, Marin-kinetics-2021, Bartholomew-fundamentals-2011, Dumesic-microkinetics-1993}, as the surface of a catalyst at which the chemical transformations of interest occur acts as a ’microdomain’.
The rate constants and other parameters for these equations are estimated based on experimental data, empirical correlations, or ab initio modelling.

In this work, we adhere to the set of elementary reactions suggested in \cite{Todic-CO-insertion-2014,Todic-corrigendum-2015}.
This model follows CO insertion mechanism of FTS and considers the stationary state of the reacting system, when a dynamic equilibrium has been established between the rates at which different intermediate substances are produced and consumed.
As a result, the balance constraints lead to a non-linear algebraic equation which needs to be solved in order to find the dependence of the overall transformation on the thermodynamic conditions {\cite{Todic-CO-insertion-2014,Todic-corrigendum-2015}}:
%
%
%
%
%
%
%
%
\begin{equation}
  \frac{1}{[S]} = c_0 + c_S \cdummy \alpha_1 \cdummy (1 + \alpha_2 + \alpha_2
  \alpha_3 + \alpha_2 \alpha_3 \alpha_4 + \cdots) 
  \label{S-equation-c0-cS}
\end{equation}
(where the summation continues up to infinity and its properties will be discussed in more details in subsection \ref{subsection-asympt-an}).
Here, the fraction of vacant catalytic sites $[S]$ is the unknown which determines the overall reaction rates and is an implicit function of the coefficients
\begin{equation}
  c_0 = 1 + K_1 P_{\mathrm{CO}} + \sqrt{K_2 P_{\mathrm{H}_2}} \hspace{0.27em},
  \label{eq:c_0}
\end{equation}
\begin{equation}
  c_S = \frac{1}{K_2^2 K_4 K_5 K_6}  \frac{P_{\mathrm{H}_2
  \mathrm{O}}}{P_{\mathrm{H}_2}^2} + \sqrt{K_2 P_{\mathrm{H}_2}}
  \hspace{0.27em}, \label{eq:c_S}
\end{equation}
These coefficients, in turn, depend on the equilibrium constants 
$$
K_j = A_j \cdummy e^{- \Delta H_j / (R T)}
$$
as the properties of the elementary steps suggested by the CO insertion mechanism as well as thermodynamic conditions at the given point of the system.
For convenience and for the sake of consistency with {\cite{Todic-CO-insertion-2014,Todic-corrigendum-2015}}, we use partial pressures $P_{CO}$, $P_{H_2}$ and $P_{H_2 O}$ of $CO$, $H_2$ and $H_2 O$ respectively, as well as the temperature $T$, to describe the thermodynamic conditions at given point.
Eq. \eqref{S-equation-c0-cS} also involves the chain growth probabilities $\alpha_j$ \cite{Sonal-recent-2021, Mendez-kinetic-2020, van-der-Laan-kinetics-1999} which are considered product-dependent and can be defined as 
\begin{equation}
  \alpha_1 = \frac{\kappa_{\tmop{growth}}}{\kappa_{\tmop{growth}} +
  \kappa_{\tmop{par}, \tmop{short}}}, \label{alpha1-def-kapp}
\end{equation}
\begin{equation}
  \alpha_2 ([S]) = \frac{\kappa_{\tmop{growth}}}{\kappa_{\tmop{growth}} +
  \kappa_{\tmop{par}, \tmop{long}} + e^{- 2 c} \cdummy
  \frac{\kappa_{\tmop{ole}, \tmop{short}}}{[S]}}, \label{alpha2-def}
\end{equation}
\begin{equation}
  \alpha_n ([S]) = \frac{\kappa_{\tmop{growth}}}{\kappa_{\tmop{growth}} +
  \kappa_{\tmop{par}, \tmop{long}} + e^{- n c} \cdummy
  \frac{\kappa_{\tmop{ole}, \tmop{long}}}{[S]}}, \: n \geqslant 3.
  \label{alpha-n3-def}
\end{equation}
by introducing a number of ‘aggregated’ parameters
\begin{equation}
  \kappa_{\tmop{growth}} = k_3 K_1 P_{\mathrm{CO}} \label{kappa-growth}
\end{equation}
\begin{equation}
  \kappa_{\tmop{par}, \tmop{long}} = k_7  \sqrt{K_2 P_{\mathrm{H}_2}}
  \label{kappa-par-long}
\end{equation}
\begin{equation}
  \kappa_{\tmop{ole}, \tmop{long}} = k_{8, 0} \label{kappa-ole-long}
\end{equation}
\[ \kappa_{\tmop{par}, \tmop{short}} = k_{7 \mathrm{M}}  \sqrt{K_2
   P_{\mathrm{H}_2}}  \]
\[ \kappa_{\tmop{ole}, \tmop{short}} = k_{8 E, 0}  \]
where $k_j$ denotes the (temperature-dependent) rate constants of elementary reactions and are reported in {\cite{Todic-CO-insertion-2014,Todic-corrigendum-2015}} along with $c = \Delta E / (R T)$.
Note that in contrast to the original definitions, we define $c > 0$, which provides a more convenient notation for asymptotics analysis performed below.

When $[S]$ for the given reaction conditions has been found by solving Eq. \eqref{S-equation-c0-cS}, the production rates for the final products of FTS are obtained as {\cite{Todic-CO-insertion-2014,Todic-corrigendum-2015}}
\begin{equation}
  R_{\mathrm{C}_n \mathrm{H}_{2 n + 2}} = \kappa_{\tmop{par}, \tmop{long}}
  \cdummy \alpha_1 \alpha_2 \cdummy \ldots \cdummy \alpha_n \cdot
  [\mathrm{S}]^2, \label{eq:R_CnH2np2}
\end{equation}
\begin{equation}
  R_{\mathrm{CH}_4} = \kappa_{\tmop{par}, \tmop{short}} \cdummy \alpha_1 \cdot
  [\mathrm{S}]^2 \hspace{0.27em}, \label{eq:R_CH4}
\end{equation}
in case of ‘long’ ($n \geqslant 2$, $n$ being the number of carbon atoms in the hydrocarbon chain) and short ($n = 1$) paraffins
respectively, while for the olefins, the similar expressions for the production
rates are given by
\begin{equation}
  R_{\mathrm{C}_n \mathrm{H}_{2 n}} = \kappa_{\tmop{ole}, \tmop{long}} \cdummy
  e^{c \cdot n} \alpha_1 \alpha_2 \cdummy \ldots \cdummy \alpha_n \cdot
  [\mathrm{S}], \label{eq:R_C2H2n}
\end{equation}
\begin{equation}
  R_{\mathrm{C}_2 \mathrm{H}_4} = \kappa_{\tmop{ole}, \tmop{short}} \cdummy
  e^{c \cdot 2} \alpha_1 \alpha_2 \cdot [\mathrm{S}] \hspace{0.27em}
  \label{eq:R_C2H4}
\end{equation}
for ‘long’ ($n \geqslant 3$) and short ($n = 2$) 1-olefins respectively.

Finally, with the reaction rates obtained using (\ref{eq:R_CnH2np2})--(\ref{eq:R_C2H4}), the total consumption rates of $H_2$ and $\tmop{CO}$ reactants can then found by appropriate stoichiometric summation as

\begin{equation}
  - R_{CO} = \sum_{n = 1}^{N_{\max}} (n \cdot
  R_{\mathrm{C}_n \mathrm{H}_{2 n + 2}}) + \sum_{n = 2}^{N_{\max}} (n \cdot
  R_{\mathrm{C}_n \mathrm{H}_{2 n}}) + \delta R_{CO}
  \label{R-CO-fin-tot}
\end{equation}

\begin{equation}
  - R_{H_2} = \sum_{n = 1}^{N_{\max}} ((2 n + 1) \cdot
  R_{\mathrm{C}_n \mathrm{H}_{2 n + 2}}) + \sum_{n = 2}^{N_{\max}} (2 n \cdot
  R_{\mathrm{C}_n \mathrm{H}_{2 n}}) + \delta R_{H_2}
  \label{R-H2-fin-tot} 
\end{equation}
%
where $N_{\max} = 100$ is the maximum carbon number treated explicitly.
Such threshold is introduced solely because of the computational efficiency reasons, as formally the upper bound of all summations should have corresponded to $N_{max} = \infty$.
In order to reduce the incaccuracy related to the truncation of the summations, we thus introduced the `corrections' 
%
\begin{multline}
\delta R_{\tmop{CO}} = L (\alpha_{N_{\max} + 1}, N_{\max}) \cdot R_{C_{N_{\max}} H_{2 N_{\max} + 2}} 
\EqLineBreak
+ L (e^c \alpha_{N_{\max} + 1}, N_{\max}) \cdot R_{C_{N_{\max}} H_{2 N_{\max}}}
\end{multline}
and
\begin{multline}
\delta R_{H_2} = 2 \cdot L (\alpha_{N_{\max} + 1}, N_{\max}) \cdot R_{C_{N_{\max}} H_{2 N_{\max} + 2}} 
\EqLineBreak
+ R_{\mathrm{C}_{N_{\max}} \mathrm{H}_{2 N_{\max} + 2}} \cdot \frac{\alpha_{N_{\max} + 1}}{1 - \alpha_{N_{\max} + 1}} 
\\
+ 2 \cdot L (e^c \alpha_{N_{\max} + 1}, N_{\max}) \cdot R_{C_{N_{\max}} H_{2 N_{\max}}}
\end{multline}
defined through an auxiliary function
\begin{equation}   L (x, N_0) = \left( N_0 + \frac{1}{1 - x} \right)  \frac{x}{1 - x}   \label{Lfunc-def} \end{equation}
(see Appendix \ref{appendix-to-inf-corrs} for a detailed derivation).
%
%
%

By the similar computation based on FTS equations stoichiometry, the total enthalpy released into the system as a result of FTS reactions can be computed.
However, in the present study we ignore its effect on the temperature change of the system for the sake of simplicity, as such a simplification preserves all the key properties of the reaction-diffusion system related to our objectives.
Unless otherwise specified, all quantities analyzed in the Results section will refer to $T = 493.15 \mathrm{K}$.



Ability of the microkinetics model to find the overall consumption rates for the FTS reactants can now to be put into the context of reactor modeling.
Through its boundary, the pellet interacts with the reactor bulk flow.
Obtaining the relationship between the the thermodynamic conditions on the pellet boundary and its productivity as a sink or source in the reactor volume is thus an essential element in linking the microkinetics scale of atomic-level processes and the scale of entire reactor within the ground-up approach to the simulations. 

To this end, we focus on the scale of the catalyst pellets. 
In simplest approximation, the catalyst pellet can be modeled as a spherical particle (Fig. \ref{fig:react-diffus-schematics}) in which the transport of the reactants and products is coupled to their chemical intercoversions. 
The substances transport within the porous volume of the pellets is typically modeled as a diffusion process, with the source terms determined by the reaction rates predicted by the microkinetics model. 
The simulation of the substances distribution within the pellets is thus a multi-staged problem built on top microkinetics model reviewed above.

In order to formulate this `pellet-scale' problem, we will consider a spherical pellet
, and describe the substances transport as a steady-state diffusion governed by a Fick's law, with constant diffusivities equal to 
$
D_{\mathrm{CO}} = 1.3 \cdot 10^{-8} \cdot \frac{\varepsilon_p}{\tau_p} \, \mathrm{\frac{m^2}{s}}
$,
$
D_{\mathrm{H_2}} = 3.6 \cdot 10^{-8} \cdot \frac{\varepsilon_p}{\tau_p} \, \mathrm{\frac{m^2}{s}}
$,
$
D_{\mathrm{H_2O}} = 1.7 \cdot 10^{-8} \cdot \frac{\varepsilon_p}{\tau_p} \, \mathrm{\frac{m^2}{s}}
$, $\epsilon_p = 0.62$, $\tau_p = 2$.
As a further simplification, we also assume that only initial reactants ($H_2$ and $\tmop{CO}$) and ultimate FTS synthesis products (paraffins, olefins and $H_2 O$) participate in mass transport across the pellet, while all intermediate substances do not. 
Such assumption allows using microkinetic model in a ‘local’ or
‘pointwise’ manner, eliminating the need to solve diffusion equations for
each of the intermediates occurring in elementary transformations.

With the above assumptions, it is now possible to formulate the equations governing the distribution of substances in the pellet as
\begin{equation}
  - \frac{1}{r^2}  \frac{d}{dr}  \left( D_j \cdot r^2  \frac{dc_j}{dr} \right)
  = s_j, \label{diffusion-eq-dim}
\end{equation}
where
$r$ is the radial coordinate of the spherical coordinate system, 
$j$ enumerates the substances ($j = CO, H_2, H_2 O$), $c_j (r)$ is the concentration (specifically, the amount concentration, in mol/m${}^3$) of $j$-th substance and 
$D_j$ is its corresponding ‘effective’ diffusion coefficient (selected so that the Fick's law could adequately approximate the flux of substances in the pellet pores), 
$s_j (\{ c_k \})$ is the source term equal to the amount of the $j$-th substance produced per unit volume per unit time.
Importantly, $s_j$ depend on all concentrations $c_j$, but not on $r$ explicitly.
They are obtained as 
\begin{equation}   s_j = \rho_{cat} \cdot R_{j}  (P_{\tmop{CO}},   P_{H_2}, P_{H_2 O}, T) 
\label{non-dim-src-term}
\end{equation} 
where $\rho_{cat} = 1980 \; \mathrm{kg}/\mathrm{m}^3$ is the effective density of the catalyst material. 
Dependence of $R_j$ on the partial pressures of substances is determined by  \eqref{R-CO-fin-tot}, \eqref{R-H2-fin-tot}, with the proper conversion of units.

Additional clarification needs to be made here regarding the arguments on which the source terms in \eqref{non-dim-src-term} depend.
Although original microkinetic model {\cite{Todic-CO-insertion-2014,Todic-corrigendum-2015}} was parameterized using the partial pressures of reactants as its input parameters, Eqs. (\ref{diffusion-eq-dim}) are formulated in terms of concentrations.
The conversion thus should be made between these two quantities.
To this end, it is essential to note that underlying experimental data in
{\cite{Todic-CO-insertion-2014,Todic-corrigendum-2015}} was collected using the slurry reactor, in which the mixture of reactants gases was bubbled through a liquid in which the catalyst is suspended in form of rather small ($<$ 90 {\textmu}m) solid particles.
By such design, the influence the physical transport resistances were minimized to allow for intrinsic kinetic measurements {\cite{Todic-CO-insertion-2014,Todic-corrigendum-2015}}.
However, in spite of using partial pressures in the original model, it is usually assumed that the reaction kinetics is determined by concentrations rather than pressures {\cite{frey_units_2023}}. Accordingly, the partial pressures $P_{\tmop{CO}}, P_{H_2}, P_{H_2 O}$ should rather be treated merely as the ‘proxies’ to the concentrations of corresponding substances inside of the catalyst particles.
The role of these ‘proxies’ can be formalized by taking into account that when the certain substance present in both gas phase (outside of the catalyst) and a condensed phase (e.g., in wax covering the catalyst surface) regions sharing a common boundary and being in thermodynamic equilibrium, the partial pressure of the substance in the gas phase is related to its concentration in the condensed phase by Henry's law \cite{sander_henrys_2022}.
We thus assume that Henry's law can also be used in the ‘inverse’ manner when it comes to applying the model parameterized in {\cite{Todic-CO-insertion-2014,Todic-corrigendum-2015}} to inner space of the catalyst pellets.
Specifically, given some concentrations $c_j$ of $\tmop{CO}$, $H_2$ and $H_2 O$ at certain point $r$ of the pellet, they can be converted through the Henry constant $H^{\tmop{cp}}_j $ as
\begin{equation}
  P_j(r) = c_j (r) \cdummy H^{\tmop{cp}}_j 
  \label{cj-to-Pj-TDC}
\end{equation}
into ‘effective’ partial pressures $P_j$ which these substances would have in the gas phase if they were supplied to the slurry reactor used by the authors of {\cite{Todic-CO-insertion-2014,Todic-corrigendum-2015}}.
Such pressures should then suitable to recreate the same conditions inside of the particle of the slurry reactor as the ones observed at the given point $r$ of the pellet.
In other words, when (\ref{cj-to-Pj-TDC}) is substituted into
(\ref{eq:c_S})--(\ref{eq:R_CnH2np2}), the input variables of the original microkinetics model are changed from partial pressures to concentrations, as desired.
Here, we implicitly assume that the same values of the Henry's constants are suitable in both (\ref{cj-to-Pj-TDC}) (which describes the catalytic particles of the slurry reactor used in {\cite{Todic-CO-insertion-2014,Todic-corrigendum-2015}}) and (\ref{cj-boundary-R}) (which is related to the material of the pellet being
modelled). 
The following values were used \cite{marano_characterization_1997, mandic_effects_2017}:
$H^{\tmop{cp}}_{\tmop{CO}} = 0.165 \tmop{bar}  \cdummy \frac{\mathrm{m}^3}{\tmop{mol}}
$, 
$
H^{\tmop{cp}}_{\textrm{H\tmrsub{2}}} = 0.222 \tmop{bar} \cdummy  \frac{\mathrm{m}^3}{\tmop{mol}}
$, 
$
H^{\tmop{cp}}_{\textrm{H\tmrsub{2}O}} = 0.0291  \tmop{bar} \cdummy   \frac{\mathrm{m}^3}{\tmop{mol}}
$.
For the sake of simplicity, all Henry's constants are assumed to be temperature- and concentrations-independent.

Together with  \eqref{R-CO-fin-tot}, \eqref{R-H2-fin-tot}, Eqs.  \eqref{cj-to-Pj-TDC}, \eqref{non-dim-src-term} and \eqref{diffusion-eq-dim} form the system of interrelated non-linear differential equations. 
By solving this system, the concentration ‘profiles’ $c_j (r)$ of the pellet can be found and these data can then be used for further analysis either on its own, or be converted into the volumetric production or consumption rates of all substances and used in a larger-scale model of FTS reactor.

Reaction-diffusion system \eqref{diffusion-eq-dim} requires also the set boundary conditions to be defined.
Similarly to {\cite{wang_modeling_2001,ghouri_multi-scale_2016,todic_effects_2018}}, no-flux boundary conditions ($\left. \frac{d c_j}{d r} \right|_{r = 0} = 0$) are used for all concentrations at the center of the pellet.
On the outer surface of the pellet, the boundary conditions for the concentrations $
c_j^{\tmop{surf}} = c_j (r = R_p)
$ 
(we used $R_p = 0.85 \; \mathrm{mm}$ throughout this work)
are obtained from Henry's law as
\begin{equation}
  c_j (r = R_p) = \frac{P_j^{\tmop{surf}}}{H^{\tmop{cp}}_j}
  \label{cj-boundary-R}
\end{equation}
where $H^{\tmop{cp}}_j$ is the Henry's constant for the $j$-th substance and
$P_j^{\tmop{surf}}$ is equal to its partial pressures in the medium surrounding the pellet, which is assumed to be in a gas phase found in the reactor volume.
The possible presence of liquid film on the surface of the catalyst pellet is neglected in our current model.

For the actual computations, the system of reaction-diffusion equations (\ref{diffusion-eq-dim}) as well as their boundary conditions were non-dimensionalized by introducing a `reference concentration'
$\ c_{\tmop{ref}} = \frac{P_{\max}}{R T_{\min}} = 1526.5
   \frac{\tmop{mol}}{\mathrm{m}^3} $
(the amount concentration observed in an ideal gas at $P_{\max} = 6
\tmop{MPa}$).
As we expect that the concentration of reactants inside of the pellet is smaller than that in its surrounding gas phase, the introduced estimate should provide a reasonable upper bound for the concentrations observed in the pellet.
Accordingly, non-dimensionalized concentrations are introduced as
$$
w_j = \frac{c_j}{c_{\tmop{ref}}} .
$$
Finally, using the pellet radius as the unit of length, we consider 
$x = \frac{r}{R_p}$
as a new non-dimensionalized radial coordinate.
In these non-dimensionalized quantities, the Eqs.(\ref{diffusion-eq-dim}) read
\begin{equation}
  - \frac{1}{x^2} \frac{d}{dx}  \left( x^2 \cdot \frac{d w_j}{d x} \right) =
  \widetilde{s_j} (x), 
  \label{wj-eq-final}
\end{equation}
where 
$
\widetilde{s_j} (x) =   \frac{R_p^2}{D_j c_{\tmop{ref}, j}} \cdummy s_j  (\{ c_k (x R_p) \}, T ) 
$
is the non-dimensionalized source term.

Note that stoichiometry of \eqref{gross-paraffin-reaction-eq}, \eqref{gross-olefin-reaction-eq} suggests that the molar production rate of $\mathrm{H_2 O}$ equals the molar consumption rate of CO: $R_{H_2 O} = - R_{CO}$ (CO is the only source of oxygen for $\mathrm{H_2 O}$ being produced as one of FTS products).
This allows effectively eliminating one of the unknowns, viz. $c_{H_2 O}$, from the system.
Indeed, it can be directly verified that as soon as there is some function $w_{CO}(x)$  satisfying \eqref{wj-eq-final} with $j=\mathrm{CO}$ and its boundary conditions,
\begin{equation}
  w_{\textrm{H\tmrsub{2}O}} (x) = - \;
  \frac{D_{\tmop{CO}}}{D_{\textrm{H\tmrsub{2}O}}} w_{\tmop{CO}} (x) + \left(
  \frac{D_{\tmop{CO}}}{D_{\textrm{H\tmrsub{2}O}}} \right)_{x = 1}
  w_{\tmop{CO}} (1) + w_{\textrm{H\tmrsub{2}O}} (1)  
  \label{w-H2O-via-w-CO} 
\end{equation}
will satisfy \eqref{wj-eq-final} with $j=\mathrm{H_2 O}$ as well as its boundary conditions.
Here, $w_j (1)$, $j = \tmop{CO} \text{ or } H_2 O$, represents the non-dimensionalized concentrations on the outer surface of the pellet, determined by the boundary conditions.
It is thus sufficient to solve \eqref{wj-eq-final} for the concentrations of $\mathrm{H_2}$ and CO only, while using applying \eqref{w-H2O-via-w-CO} to eliminate $\mathrm{H_2 O}$ concentration from their source terms.
%
%

As can be seen from the above discussion, even within a simplified model, the ground-up modeling of the distributions of substances inside of the catalyst pellet not only requires numerically solving diffusion equations, but also computing their source terms at each iteration of the solver.
This further requires entire microkinetics model to be simulated `on-the-fly', including solving of non-linear equation \eqref{S-equation-c0-cS}. It is thus advantageous to use numreical approximation methods to accelerate the computationally demanding steps by avoiding the need of explicitly solving all underlying chemical kinetics equations on each iteration.

\begin{figure*}
  \raisebox{0.0\height}{\includegraphics[trim={0 0 0 10},clip,width=\linewidth]{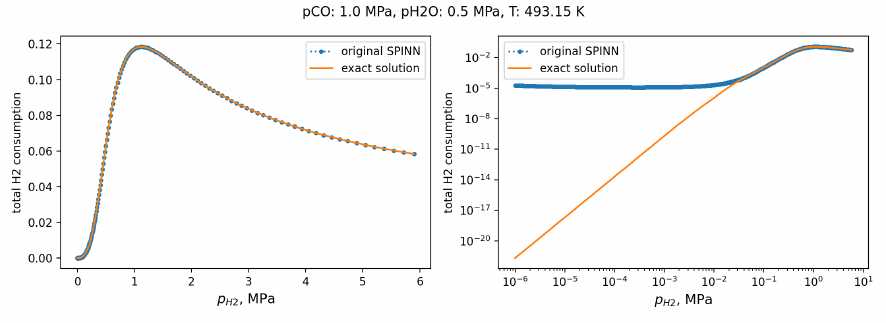}}
  \\
    \begin{tabular}{p{0.5\textwidth} p{0.5\textwidth}}
        \centering {\it a} & \centering {\it b} \\ 
    \end{tabular}

  \caption{
  $H_2$ consumption rate $- s (c)$ as a function of $H_2$ concentration $c$ (at $P_{CO} = 1 \; \mathrm{MPa}$, $P_{H_2 O} = 0.5 \; \mathrm{MPa}$, $T = 493.15 \; \mathrm{K}$). The difference between exact and approximated dependencies might seem negligible when characterized by the absolute difference (and also hardly noticeable in linear scale, \tmtextit{a}), but exhibits drastically different asymptotic behaviour (best viewed in
  logarithmic scale, \tmtextit{b}).
  \label{fig:original-SPINN-to-H2-rates}
  }
\end{figure*}

As we have previously shown in \cite{our-prev-arxiv-preprint}, neural networks, specifically PINNs, can be applied for this purpose.
Still, as we will demonstrate in subsection \ref{model-example-subsection}, special care must be taken when any approximation is used in the source terms of the reaction-diffusion equations.
In particular, we will show that seemingly minor inaccuracies (cf. Fig. \ref{fig:original-SPINN-to-H2-rates}) in the approximation of $s (c)$ can lead to drastically unphysical results, such as negative concentrations.
We will then proceed by analyzing the true asymptotic behaviour of the chosen microkinetics model, modifying original PINN model to match those asymptotics by design and demonstrating that with such a modification, the PINN model becomes applicable for accelerating the source terms of the reaction-diffusion equations.
%

\subsection{A simplified ‘toy model’ example 
}
\label{model-example-subsection} 

\begin{figure*}
  \raisebox{0.0\height}{\includegraphics[width=0.92\linewidth]{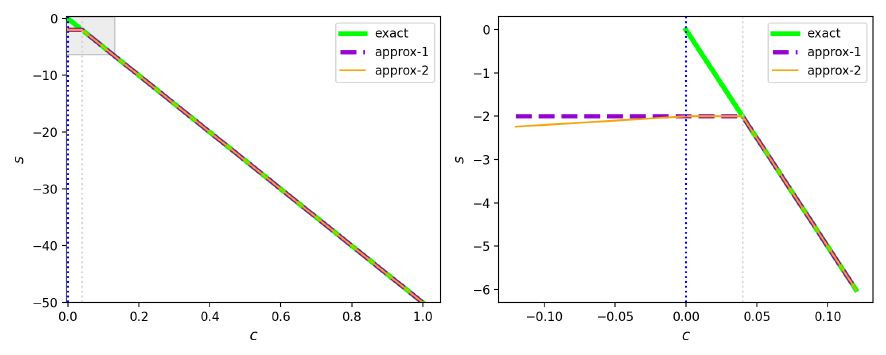}}
  \caption{\label{lin-sc-and-its-approxs}Exact dependence of the source term
  $s$ on substance concentration $c$ and two approximations which differ from
  it only in a small region (zoomed in a right plane) near $c = 0$}
\end{figure*}

In order to showcase how minor inaccuracies (or, more properly, incorrect
asymptotics) in approximation of $s (c)$ even in a small region near $c = 0$
can lead to unphysical solutions of (\ref{model-diff-1}), the following simple
model example can be helpful. Consider a simple one-dimensional
reaction-diffusion system, which can be described, in its stationary state, by
a differential equation
\begin{equation}
  \frac{d^2 c}{dx^2} = - s (c) \label{model-diff-1}
\end{equation}
with boundary conditions $\frac{dc}{dx}  (x = 0) = 0$ and $c (x = 1) = 1$ (for
convenience, we'll use nondimensionalized units throughout this example). Such
system can correspond to a transport of some substance in a reactive media,
where this substance is consumed in a chemical reaction. The dependence of the
source term $s < 0$ on the concentration $c (x)$ is then determined by a
function $s (c)$ resulting from a complicated chain of chemical
transformations. For the sake of this model example, we assume that the exact
$s (c)$ is approximated or fitted by some other function (e.g., by an NN). To
simplify the system even further, assume that the ‘exact’ $s (c)$ is given
by
\begin{equation}
  s (c) = - k \cdummy c, \label{skc-exact1}
\end{equation}
and consider two of its ‘approximations’, which differ from the exact form only in a small region near $c = 0$ (Fig. \ref{lin-sc-and-its-approxs}):
\begin{equation}
  s^{\tmop{approx}}_1 (c) = - k \cdummy \left\{ \begin{array}{ll}
    \theta, & c < \theta\\
    c, & c \geqslant \theta
  \end{array} \right. \label{s1approx-toy}
\end{equation}
and
\begin{equation}
  s^{\tmop{approx}}_2 (c) = - k \cdummy \left\{ \begin{array}{ll}
    \frac{| c |}{25}, & c < 0\\
    \theta, & 0 \leqslant c < \theta\\
    c, & c \geqslant \theta
  \end{array} \right. \label{s2-approx-toy}
\end{equation}
In all cases we set $k = 50$ and $\theta = 0.04$, and used {\texttt{fsolve}} from SciPy library \cite{virtanen_scipy_2020} to obtain the solutions numerically, althoug all three considered forms of $s (c)$ can be treated analytically as well, additionally requiring the solution of a single non-linear algebraic equation when $s_1^{\tmop{approx}}$ or $s_2^{\tmop{approx}}$ are involved.

It is essential that although the region of negative concentrations is unphysical, all the three functions are defined for that region as well.
When the exact dependence (\ref{skc-exact1}) is used, the properly constructed numerical solver will never call $s (c)$ with negative argument $c$ in given problem.
This might change however, when ‘approximate’ $s (c)$ is used in the equation instead.

\begin{figure*}[h]
  \raisebox{0.0\height}{\includegraphics[width=0.92\linewidth]{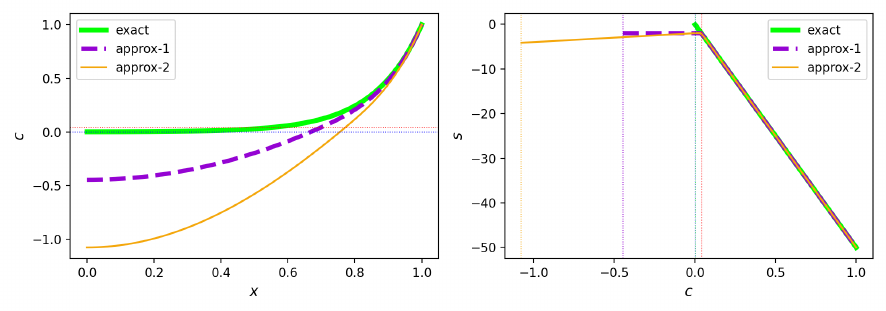}}
  \caption{\label{fig-toy-example-profiles}The concentration profiles $c (x)$
  obtained with exact form of dependence of the source term $s$ on
  concentration $c$ and two its approximations (\ref{s1approx-toy}) and
  (\ref{s2-approx-toy}). Right plane explicitly shows these approximations for
  the range of concentration values requested by the numerical solver. For
  convenience, threshold values of $c = 0$ and $c = \theta$ are shown in a
  left plane with blue and red the dotted lines respectively.}
\end{figure*}

Indeed, suppose that at some $x = x_{\tmop{thresh}}$ we have $c (x_{\tmop{thresh}}) = \theta$ and approximation $s_1^{\tmop{approx}}$ is used.
In that case, given the shape of exact solution 
$c (x) = \frac{\cosh \left( x \cdummy \sqrt{k} \right)}{\cosh \left( \sqrt{k} \right)}$
(shown in Fig. \ref{fig-toy-example-profiles}), the solver can then ‘assume’ that the value of concentration should continue decreasing as $x$ decreases further to the left from $x_{\tmop{thresh}}$.
In that case, $s_1^{\tmop{approx}}$ will evaluate to $- k \cdummy \theta$ suggesting that $c$ should decrease even further at even smaller coordinates and so on. The concentration can then reach $c = 0$ value at some finite $x = x_0$.
At that point two scenarios, both resulting from the same cause of inaccurate asymptotics of $s_1^{\tmop{approx}}$, can be considered. 
In the first scenario, the function approximating the true $s (c)$ verifies its argument to ensure that it belong to a physically meaningful domain (in our case, $c \geqslant 0$) and causes the solver failure if that verification fails. 
In contrast to that, the second scanario can be considered which would more closely resemble the case when a fully-connected NN without any ‘ad hoc' modifications had been used to approximate the true ‘complicated’ $s (c)$ function.
In such scenario, it would have ‘silently’ output ‘some’ value of $s^{\tmop{approx}}$ even if its argument $c$ is unphysical (negative), and the process of solving Eq. (\ref{model-diff-1}) numerically could continue.
As a result, a curve $c (x)$
shown in Fig. \ref{fig-toy-example-profiles} is obtained.

When crafting the approximate function $s^{\tmop{approx}} (c)$ investigated in the considered scenario of our ‘toy’ example, we thus focused on two possibilities.
For the first one, $s_1^{\tmop{approx}}$ simulates the situation when the model approximating $s (c)$ was ‘trained’ only for the values of $c$ above certain finite threshold $\theta$, and when the ‘extrapolation regime’ of the approximator works by just outputting the value of $s$ it has ‘learned’ for the ‘closest’ concentration available in the training set (in our case, $\theta$).
For the second one, $s_2^{\tmop{approx}}$ simulates a similar situation, but admits that for negative arguments within the extrapolation regime the approximator can output some ‘random’ dependence on concentration rather then producing a constant output.
We thus added a ‘small’ slope to $s^{\tmop{approx}}_2$ for $c < 0$, which might be expected to ‘slightly worsen’ results from the previous approximation.
The concentration profiles obtained under these two approximations are shown in Fig. \ref{fig-toy-example-profiles}.

Obtained profiles clearly demonstrate that even minor inaccuracies in $s (c)$ asymptotic near $c = 0$ can result in qualitatively incorrect $c (x)$ profiles.
Moreover, the drift into unphysical region can even be self-accelerating, as shown by the effect of a small additional slope
introduced in $s^{\tmop{approx}}_2$ in $c < 0$ region.

In practice, when NN is used to approximate the true $s (c)$ dependence, the model is trained on a dataset composed from the discrete set of $\{ (c_i, s_i) \}$ data-points.
Because approximate model will not capture the true $s (c)$ dependence exactly, its output will have some error for all of its inputs.
In particular, the output value $s_{\tmop{NN}} (0)$ at $c = 0$ will never equal to $s_{\tmop{exact}} (0) = 0$ exactly, unless some special modifications are introduced into the NN architecture.
In turn, this can result in approximated $s_{\tmop{NN}} (c)$ dependencies with the same features as the model examples $s^{\tmop{approx}}_1$ or even $s^{\tmop{approx}}_2$ considered above.

We thus conclude that ensuring correct asymptotic behavior of $s_{\tmop{NN}}$ within some \tmtextit{continuous region} near $c = 0$ is essential to ensure that numerical solver applied to (\ref{model-diff-1}) produces physically meaningful results.
Another requirement, which has not been encountered in the considered example, but is nevertheless equally important, is to ensure that given the approximation of $s (c)$ dependence with a correct asymptotic, the numerical solver applied to the reaction-diffusion equation should never invoke the (approximated) source term $s (c)$ function with ‘unphysical' values of its argument $c$ (which can happen in, e.g., predictor-corrector or Runge-Kutta ODE solving methods).
We will address these two issues in the subsequent subsections: asymptotic in subsection \ref{subsection-asympt-an} and adapting numerical reaction-diffusion equation solver in subsection \ref{appendix-conv-solver-initial-guess}.

\section{Results and discussion}\label{sec-results-and-discuss}

\subsection{Asymptotic behavior of CO insertion FTS model in low pressure
region
}
\label{subsection-asympt-an}

As the first step in modifying the architecture of PINN, we analyze
analytically the true asymptotics which the reaction rates have as the functions of the reactants partial pressures.
When $P_{\tmop{CO}} \rightarrow 0$, all $\alpha_j$'s tend to zero due to \eqref{kappa-growth} and, most importantly, so do all the reaction rates when computed by (\ref{eq:R_CnH2np2})--(\ref{eq:R_C2H4}).
We thus will consider below only the case when $P_{\tmop{CO}} \neq 0$ and $P_{H_2 O} \neq 0$ (as even when a completely dry reactants mixture is fed into the reactor or the catalyst pellet, some water vapour will still be produced in it as a result of FTS).
By (\ref{eq:R_CnH2np2})--(\ref{eq:R_C2H4}), this case reduces to analyzing the asymptotic behaviour of $[S]$ at small $P_{H_2}$.

To perform such an analysis, we first rewrite (\ref{S-equation-c0-cS}) as
\begin{equation}
  [S] = \frac{1}{c_0 + c_S \cdummy \alpha_1 \cdummy J} \label{eq-S-via-J}
\end{equation}
by introducing an auxiliary quantity
\begin{equation}
  J = \frac{1}{\alpha_1} \cdummy \sum_{n = 1}^{\infty} \prod_{j = 1}^n
  \alpha_j = 1 + \alpha_2 + \alpha_2 \alpha_3 + \alpha_2 \alpha_3 \alpha_4 +
  \cdots \label{eq-J-definition}
\end{equation}
Importance of this quantity stems from the fact that in case when $\alpha_j$ become close to unity, this sum can potentially tend to infinity, determining thereby the ‘smallness’ of $[S]$. 
It will, according to (\ref{eq:R_CnH2np2})--(\ref{eq:R_C2H4}), also determine the asymptotic behavior of the reaction rates under the thermodynamic conditions when these
rates are ‘small’.

Apart from potentially ‘large’ $J$, $[S]$ in (\ref{eq-S-via-J}) also depends on $c_0$, which doesn't have peculiarities at small $P_{H_2}$ or $P_{\tmop{CO}}$ (and just tends to unity according to (\ref{eq:c_0})), and on $c_S$, defined by (\ref{eq:c_S}).
In contrast to $c_0$, the latter can become quite ‘large’, with an asymptotics $c_S \sim \frac{1}{P_{H_2}^2}$ when $P_{H_2} \rightarrow 0$ and when $P_{H_2 O}$ is finite and non-zero (which is quite a realistic assumption especially near the center of the pellet, where the concentration of $H_2 O$ is the largest).
It is thus 
$c_S \cdummy \alpha_1 \cdummy J$
term in (\ref{eq-S-via-J}), which can potentially become large and thus determine ‘smallness’ of $[S]$.
Even with finite $\alpha_1 J$, we can thus expect that at small $P_{H_2}$ the following estimate holds
\begin{equation}
  \frac{1}{[S]} \sim c_S \Longrightarrow [S] \sim P_{H_2}^2 .
  \label{eq-S-PH2-sq}
\end{equation}
This would, in turn, imply that all production rates (\ref{eq:R_CnH2np2})--(\ref{eq:R_C2H4}) tend to zero as $P_{H_2}$ to the power of at least 2.
Apart from its importance for the analysis of asymptotics and PINN architecture modifications, these heuristics would further imply that the overall reaction rate to concentration ratio $\frac{s}{c}$ remains finite. 
Such property appears to be an important factor in the design of a numerical solver
for the reaction-diffusion equation, as discussed in more details in section \ref{appendix-conv-solver-initial-guess}.

So far we have only assumed that $J$ remains finite rather then goes to infinity at small $P_{H_2}$, which lead to \eqref{eq-S-PH2-sq}. 
However, if $J$ as the sum of the series (\ref{eq-J-definition}) diverges, the exponent in \eqref{eq-S-PH2-sq} will be different. 
We thus proceed with investigating the actual behaviour of the sum $\alpha_1 \cdummy J$ more carefully at small $P_{H_2}$.
To this end, consider $\alpha_n$ from (\ref{alpha-n3-def}) with $n \geqslant 3$:
\begin{equation}
  \alpha_n = \frac{1}{1 + \frac{\kappa_{\tmop{par},
  \tmop{long}}}{\kappa_{\tmop{growth}}} + \frac{\kappa_{\tmop{ole},
  \tmop{long}}}{\kappa_{\tmop{growth}}} \cdummy \frac{1}{[S]} \cdummy e^{- n
  c}} = \frac{1}{1 + \varepsilon + A \cdummy e^{- n c}}
  \label{eq-alpha-n-via-eps-A}
\end{equation}
where we introduced a ‘smallness parameter’
\begin{equation}
  \varepsilon = \frac{\kappa_{\tmop{par},
  \tmop{long}}}{\kappa_{\tmop{growth}}} = \frac{K_7  \sqrt{K_2}}{K_3
   K_1 } \cdot {\frac{\sqrt{P_{H_2}}}{P_{\tmop{CO}}}}
  \label{eq-epsilon-kappa-ratio-def}
\end{equation}
and 
$A = \frac{1}{[S]} \frac{\kappa_{\tmop{ole},
\tmop{long}}}{\kappa_{\tmop{growth}}} = \frac{1}{[S]} \frac{K_{8, 0}}{K_3
\cdot K_1 P_{\tmop{CO}}}$ 
for a shorthand notation.

\begin{proposition}
$J$ remains finite as $\varepsilon$ goes to zero.
\end{proposition}

\begin{proof}
In order for $J$ to remain finite, the $\alpha_2 \cdummy \ldots \cdummy \alpha_n$ products in (\ref{eq-J-definition}) should tend to zero as $n$ rises.
This is possible due to $\alpha_j < 1$ inequality which, in turn, results from the presence of both $\varepsilon$ and $A \cdummy e^{- c n}$ in (\ref{eq-alpha-n-via-eps-A}) at ‘small’ $n$, but both of them can vanish leading to $\alpha_j \approx 1$ at ‘large’ $n$, as $A \cdummy e^{- c n}$ decays and in case $\varepsilon \rightarrow 0$ limit is considered. 
Indeed, the presence of $\varepsilon$ alone in the denominator of (\ref{eq-alpha-n-via-eps-A}) would result in
$\alpha_n \approx \frac{1}{1 + \varepsilon} = \alpha_{\infty}$ 
(which is assumed for $n$ as low as $n
\geqslant 2$, for the sake of simplicity), and we would end up with having
\begin{equation}
  J \approx 1 + \alpha_{\infty} + \alpha_{\infty}^2 + \cdots = \frac{1}{1 -
  \alpha_{\infty}} = \frac{1 + \varepsilon}{\varepsilon} \sim
  \frac{1}{\varepsilon}, \label{alpha-inf-sum-1-over-eps}
\end{equation}
which tends to infinity at small $\varepsilon$.
As the ‘diverging’ part of this sum is dominated by the large-index terms, the products 
$\alpha_2 \cdummy \ldots \cdummy \alpha_n$
in (\ref{eq-J-definition}) with ‘large’ $n \gg 1$ can be analyzed first. 
In that regime, there is a ‘competition’ of two multipliers. 
One of them, which ultimately can be factored out in (\ref{eq-J-definition}), comes from some number $N$ of ‘initial’
$\alpha_j$'s in which the presence of non-negligible
$A \cdummy e^{- c j} \geqslant A \cdummy e^{- c N} > \varepsilon$
results in 
$\alpha_2 \cdummy \ldots \cdummy \alpha_N \ll 1$
being ‘small’, while the second multiplier corresponds to the remaining
$\alpha_{N + 1} \ldots \alpha_n \approx \alpha_{\infty}^{n - N}$ 
with a \eqref{alpha-inf-sum-1-over-eps}-like divergence at small $\varepsilon$:
\begin{multline}
  J \sim 1 + \cdots + \alpha_2 \cdummy \ldots \cdummy \alpha_N \cdummy (1 +
  \alpha_{N + 1} + \alpha_{N + 1} \alpha_{N + 2} + \cdots)  
  \EqLineBreak
  \sim \alpha_2
  \cdummy \ldots \cdummy \alpha_N \cdummy \frac{1}{\varepsilon}
  \label{eq-J-asympt-factored-out-1}
\end{multline}
In other words, it should be checked whether at small $\varepsilon$ the factored-out part $\alpha_2 \cdummy \ldots \cdummy \alpha_N$ can under small $\varepsilon$ go to zero faster than $\sim \frac{1}{\varepsilon}$ sum of the remaining parts rises.
In such analysis, we should treat $A \sim \frac{1}{[S]}$ as a relatively ‘large’ number, because whether $J$ remains finite or not, $[S] \approx \frac{1}{c_S \cdummy \alpha_1 \cdummy J}$ is guaranteed to be small at small $\varepsilon$, at least, because of $c_S \sim \frac{1}{\varepsilon^4}$ at small $\varepsilon$ due to (\ref{eq:c_S}).
More
strictly, we have
\begin{multline} [S] = \frac{1}{c_0 + c_S \alpha_1 J} < \frac{1}{\alpha_1 c_S J} 
\\ <
   \frac{1}{\alpha_1 \left( \frac{1}{K_2^2 K_4 K_5 K_6}  \frac{P_{\mathrm{H}_2
   \mathrm{O}}}{P_{\mathrm{H}_2}^2} + \sqrt{K_2 P_{\mathrm{H}_2}} \right)} <
   \frac{1}{\alpha_1 \left( \frac{1}{K_2^2 K_4 K_5 K_6}  \frac{P_{\mathrm{H}_2
   \mathrm{O}}}{P_{\mathrm{H}_2}^2} \right)} 
  \EqLineBreak
   = \frac{1}{b} \cdummy
   \varepsilon^4
\end{multline}
where we used (\ref{eq:c_S}), relied on $J \geqslant 1$ and introduced 
$b = \alpha_1 \left( \frac{P_{\mathrm{H}_2 \mathrm{O}}}{K_2^2 K_4 K_5 K_6} \right) \left( \frac{K_7 \cdot \sqrt{K_2}}{K_3 \cdot K_1 } \frac{1}{P_{\tmop{CO}}} \right)^4$.
This implies $[S] < \frac{1}{b} \cdummy \varepsilon^4 \Rightarrow \frac{1}{[S]} > \frac{b}{\varepsilon^4}$, so that
\begin{multline}
  \alpha_j = \frac{1}{1 + \varepsilon + \frac{1}{[S]}
  \frac{\kappa_{\tmop{ole}, \tmop{long}}}{\kappa_{\tmop{growth}}} \cdummy e^{-
  c j}} < \frac{1}{1 + \varepsilon + \frac{b}{\varepsilon^4}
  \frac{\kappa_{\tmop{ole}, \tmop{long}}}{\kappa_{\tmop{growth}}} \cdummy e^{-
  c j}} 
  \EqLineBreak
  = \frac{\varepsilon^4}{\varepsilon^4 + \varepsilon^5 + b
  \frac{\kappa_{\tmop{ole}, \tmop{long}}}{\kappa_{\tmop{growth}}} \cdummy e^{-
  c j}} \label{eq-alphaj-with-large-A}
\end{multline}
Although it can be seen that for sufficiently large $j$, when 
$b \frac{\kappa_{\tmop{ole}, \tmop{long}}}{\kappa_{\tmop{growth}}} \cdummy e^{- c j} \ll \varepsilon^5$,
this upper bound tends to the same
$\frac{1}{1 + \varepsilon} = \alpha_{\infty}$
as in \eqref{alpha-inf-sum-1-over-eps}, for smaller $j$'s (which are of interest in \eqref{eq-J-asympt-factored-out-1}), when 
$b \frac{\kappa_{\tmop{ole}, \tmop{long}}}{\kappa_{\tmop{growth}}} \cdummy e^{- c j} \gg \varepsilon^4$,
each $\alpha_j$ is bounded from above by a quantity $\sim \varepsilon^4$.
The product $\alpha_2 \cdummy \ldots \cdummy \alpha_N$ in (\ref{eq-J-asympt-factored-out-1}) is thus less than $\varepsilon$ to the power higher than 1, so that $J$ is guaranteed to remain finite at small $\varepsilon$.
\end{proof}

\begin{figure*}
  \raisebox{0.0\height}{\includegraphics[width=\linewidth]{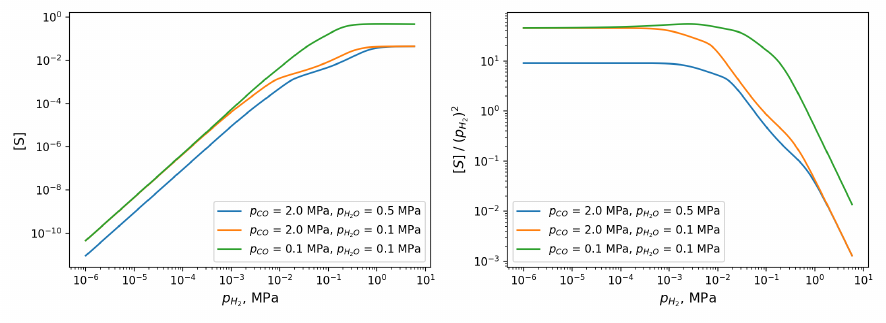}}
  \caption{\label{fig-logS-logPH2-2}Dependence of $[S]$ on $P_{H_2}$ at $T =
  493.15 K$ under different values of $P_{\tmop{CO}}$ and $P_{H_2 O}$ obtained
  using numerical equation solver for \eqref{S-equation-c0-cS} with
  microkinetics parameters taken from
  {\cite{Todic-CO-insertion-2014,Todic-corrigendum-2015}}. Right panel
  additionally shows the ratio $\frac{[S]}{(P_{H_2})^2}$ for convenience.}
\end{figure*}

We have thus shown that when $\varepsilon \sim
\frac{\sqrt{P_{H_2}}}{P_{\tmop{CO}}}$ is made small enough, $J$ remains
finite.
This further means that in $c_S \alpha_1 \cdummy (1 + \alpha_2 +
\alpha_2 \alpha_3 + \alpha_2 \alpha_3 \alpha_4 + \cdots)$ the term $c_S
\cdummy \alpha_1$ will dominate, so that \eqref{eq-S-PH2-sq}, initially
introduced as a heuristics for the asymptotics of $[S]$ at small
$\varepsilon$, is now verified. 
%
This can be also confirmed by an additional
numerical verification as shown in Fig. \ref{fig-logS-logPH2-2}, where
$\frac{[S]}{(P_{H_2})^2}$ ratio tends to a constant at small $P_{H_2}$.

As will be discussed in the next subsection, the performed analysis is also
useful for designing a post-processing transformation needed to ensure that
the dependence of $[S]$ on the reaction conditions learned by the PINN is
consistent with the obtained asymptotics at small $P_{H_2}$.

\subsection{Post-processing transformation for SPINN output
}
\label{subsection-SPINN-post-processing}

It is known that vanishingly small concentrations of the FTS reactants are quite common in the inner region of the pellet. At the same time, most, if not all, of the data points which are fed into SPINN during its training process correspond to the partial pressures of the reactants which are noticeably non-zero due to pseudo-random sampling of training points.
In other words, unless some non-standard training protocol is employed, the region of the SPINN input space which is of interest during the inference phase (e.g., when SPINN is invoked as a part of the diffusion equation solver) is broader than the one used during the training.
It is thus critical to ensure that SPINN outputs exhibit correct asymptotic behavior as the functions of partial pressures, when $H_2$ or $\tmop{CO}$ partial pressures become small, i.e., when the inference is performed in an extrapolation regime.

In our approach, this requirement is fullfilled by introducing an \tmtextit{ad hoc} output layer, which we refer to as the post-processing transformation $G$. This transformation relates the ‘raw output’ of SPINN (implemented, e.g., as a fully-connected network with commonly used activation functions) to the ultimate quantity of interest $[S]$ as
$$
[S] = G (\tmop{SPINN} (\bar{X})) = G(y), 
$$
where $y$ is a `raw' NN output, and
\begin{equation}
  \bar{X} = \left\{ \frac{P_{\mathrm{CO}}}{P_{\mathrm{CO}}^{\max}},
  \frac{P_{\mathrm{H}_2}}{P_{\mathrm{H}_2}^{\max}}, \frac{P_{\mathrm{H}_2
  \mathrm{O}}}{P_{\mathrm{H}_2 \mathrm{O}}^{\max}}, \frac{T -
  T^{\min}}{T^{\max} - T^{\min}} \right\} \label{eq-Xbar-inputs-def}
\end{equation}
denotes the non-dimensionalized thermodynamic conditions which are fed into SPINN as its inputs.
Here, we used
\newcommand{\TrainCondRanges}{
$P_{\mathrm{CO}}^{\max} = P_{\mathrm{H}_2}^{\max} = 6.0 \; \mathrm{MPa}$,
$P_{\mathrm{H}_2 \mathrm{O}}^{\max} = 6.1 \; \mathrm{MPa}$,
$T^{\min} = 473.15 \; \mathrm{K}$,
$T^{\max} = 513.15 \; \mathrm{K}$
}
\TrainCondRanges
as the ranges which reasonably reflect the thermodynamic conditions found in FTS chemical reactors (cf. \cite{our-prev-arxiv-preprint}).

Based on the heuristics discussed in Appendix \ref{post-processing-heuristics-appendix}, we constructed the post-processing transformation as a two-stage procedure.
First, we introduce a `naive' approximation for $\alpha_2$ as
\begin{equation}
  \tilde{\alpha}_2 = (\alpha_2)_{[S] = \sigma} =
  \frac{\kappa_{\tmop{growth}}}{\kappa_{\tmop{growth}} + \kappa_{\tmop{par},
  \tmop{long}} + e^{- 2 c} \cdummy \frac{\kappa_{\tmop{ole},
  \tmop{short}}}{\sigma}}
  \label{}
\end{equation}
where $
  \sigma (y) = \left( c_0 + c_S \cdummy \alpha_1 \cdummy \left( 1 +
  \frac{\alpha_{\infty}}{1 - \alpha_{\infty}} \cdummy y \right)
  \right)^{-1}
$ is a `naive' counterpart of $[S]$ (cf. \eqref{sigma-y}) and $\alpha_{\infty} = \frac{1}{1 + \varepsilon}$.
We then use it to relate the actual target quantity $[S]$ to the NN output $y$ as
\begin{equation}
  [S] = G (y) = \frac{1}{c_0 + c_S \cdummy \alpha_1 \cdummy \left( 1 +
  \tilde{\alpha}_2 + \tilde{\alpha}_2 \cdummy \frac{\alpha_{\infty}}{1 -
  \alpha_{\infty}} \cdummy y \right)} \label{Gy-final-text}
\end{equation}
When $\sigma$ is small, 
$\tilde{\alpha}_2 \approx \sigma \frac{\kappa_{\tmop{growth}}}{e^{- 2 c} \cdummy \kappa_{\tmop{ole}, \tmop{short}}} \sim \varepsilon^5$, 
which is small enough to ‘damp’
$\frac{\alpha_{\infty}}{1 - \alpha_{\infty}} \cdummy y \sim \frac{1}{\varepsilon}$, ensuring that $1 + \tilde{\alpha}_2 + \tilde{\alpha}_2 \cdummy \frac{\alpha_{\infty}}{1 - \alpha_{\infty}} \cdummy y \approx 1$ 
for small $\varepsilon$.
Accordingly, the dependence of $[S]$ on $\varepsilon$ becomes
determined by $c_S \sim {P_{H_2}}^{-2}$ and thus in fact independent of $y$, as required.
When $\varepsilon$ is not as small (say, at the boundary of the interpolation region of SPINN inputs), $\sigma$ can be expected to be a good guess for $[S]$. 
This is because 
$1 + \frac{\alpha_{\infty}}{1 - \alpha_{\infty}} \cdummy y = 1 + \frac{y}{\varepsilon} \approx 1$ 
is $\sigma (y)$ as soon as $\varepsilon$ becomes larger than a plateau on which SPINN output $y$ rests at small input pressures, and, as a result, $\sigma$ follows the correct $\sim \varepsilon^4$ (as determined by
$c\tmrsub{S}{\sim}{\frac{1}{P\tmrsub{H\tmrsub{2}}\tmrsup{2}}}{\sim}{\frac{1}{{\varepsilon}\tmrsup{4}}})$
asymptotics, as opposed to $\sigma \sim \varepsilon^5$ which took place at smaller $\varepsilon$. 
In turn, when $\sigma \approx [S]$, $\tilde{\alpha}_2 \approx \alpha_2$ can be anticipated as well, so that $\frac{\alpha_{\infty}}{1 - \alpha_{\infty}} \cdot y$ now needs to approximate only $\alpha_3 + \alpha_3 \alpha_4 + \cdots$, effectively requiring NN to learn only the properties of \eqref{alpha-n3-def}, and not of \eqref{alpha2-def} (cd. Appendix \ref{post-processing-heuristics-appendix}).
This way, (\ref{Gy-final-text}) can properly account for the difference between $\kappa_{\tmop{ole}, \tmop{short}}$ and $\kappa_{\tmop{ole}, \tmop{long}}$.

Importantly, in \eqref{Gy-final-text} and in $\sigma(y)$ it has been implicitly assumed that SPINN is built so that $y \geqslant 0$ is ensured by design for arbitrary input $\bar{X}$, e.g., the final layer in SPINN is formed by a single ReLU function.
Such treatment restricts the choice of the applicable activation functions however. Therefore, in practice we used
\begin{equation}
  y = \tmop{ReLU} (\tmop{SPINN} (\bar{X})), \label{eq-ReLU-SPINN-clipping}
\end{equation}
where the output layer of the fully-connected neural network $\tmop{SPINN} (\bar{X})$ is a linear transformation, to which the `final' ReLU acts as an activation function.

\subsection{SPINN training and validation}
\label{train-val-subsection}

SPINN with the post-processing transformation introduced in subsection \ref{subsection-SPINN-post-processing} was implemented and trained using NVIDIA Modulus framework. 
The $\tmop{SPINN} (\bar{X})$ in \eqref{eq-ReLU-SPINN-clipping} was represented by two-layer feed forward fully connected architecture, and a single ReLU function was applied to its output according to \eqref{eq-ReLU-SPINN-clipping} to ensure that $y$ is not
negative.
Each layer contained 512 neurons and GELU activation functions \cite{hendrycks_gaussian_2023} (see also \cite{de_wolff_towards_2021, lee_mathematical_2023, richert_layered_2023, tsuchida_avoiding_2021} discussing the relevant properties of this activation function).
The network was trained for $1 \cdummy 10^6$ epochs using Adam optimizer with initial learning rate of $\eta_0 = 1 \cdummy 10^{- 3}$ and the so-called `inverse time decay' learning rate scheduler, prescribing
$
\eta = {\eta_0} \cdot \left( {1 + \kappa \cdot \frac{t}{\tau_s}} \right) ^{-1}, 
$
as the dependence of learning rate $\eta$ on the number of epoch $t$, with $\tau_s = 100$ and $\kappa = 0.01$ being the chosen values of the scheduler adjustable parameters.
The optimizer used the difference between l.h.s. and r.h.s. of (\ref{S-equation-c0-cS}) as the loss function, aggregating these values over a batch of 10\,000 input points and minigizing mean absolute error (MAE) statistics. 
The mentioned choice of hyperparameters resembled the training regime previously used to train a somewhat similar PINN in \cite{our-prev-arxiv-preprint}.

During the training, each of the inputs, as defined by
\eqref{eq-Xbar-inputs-def}, was sampled uniformly from 0{\textdots}1 range.
For both training and testing stages, these intervals of non-dimensionalized PINN input values corresponded to \TrainCondRanges in physical units.
Within the same ranges, 18 equally spaced values ($k/19$, $k=1...18$) for each of the input quantities have been further selected to build a testing dataset containing $18^4 = 104976$ points in total.
This dataset, augmented with the ground truth solutions of (\ref{S-equation-c0-cS})--(\ref{alpha-n3-def}) found by a conventional numerical solver, was used to assess the model accuracy, and contained the same set of points as the one used previously in \cite{our-prev-arxiv-preprint}.
Importantly, the lowest values of $H_2$, CO and $H_2 O$ pressures found in it were $\approx 0.32$ MPa ($\frac{1}{19} \approx 0.0526$ in non-dimensionalized units), so that although such testing dataset properly reflects the overall performance of the model in the range of thermodynamic conditions found in inter-pellet space of FTS reactors, it does not assess the models' accuracy in the asymptotic region of low pressures.
We thus refer to this dataset as a `general' one, and below report the accuracy metrics obtained with it separately from those obtained in the low-pressure regions.

As a baseline, we used the PINN previously proposed in \cite{our-prev-arxiv-preprint}, which contained 2 hidden layers, 512 neurons each, and used GELU as an activation function.
All other hyperparameters were the same as given above.
In the baseline case, 
\begin{equation}
    G(y) = 10^{y}
    \label{eq:original-post-proc-transfrm}
\end{equation}
was used as a post-processing transformation, which was thus the only distinction between it and the PINN model proposed in this work.

\begin{figure}[t]
\centering
{
\halfpagefgraphics[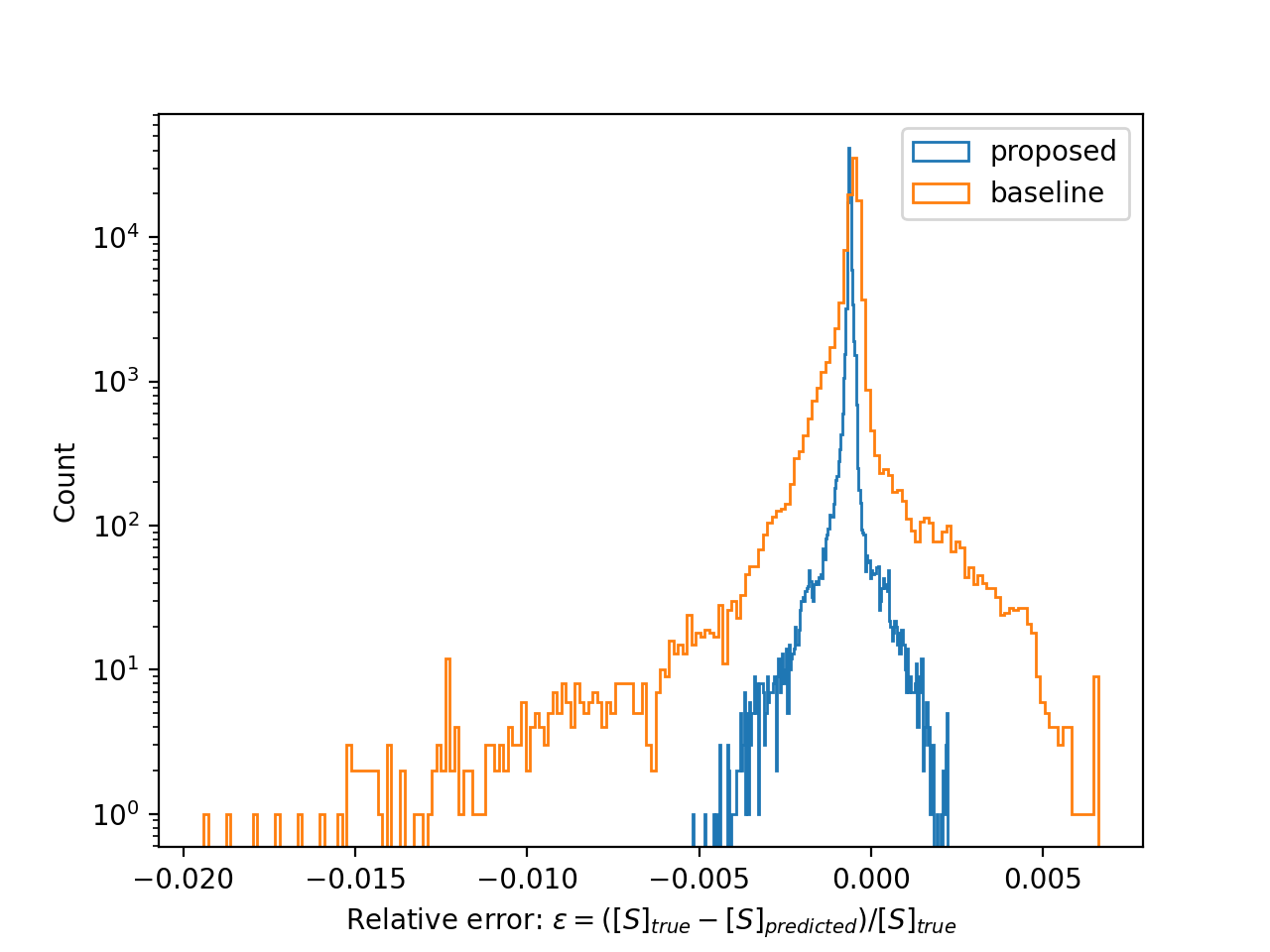]
}
  \caption{\label{fig:err_distr_gen_ds}Comparison of the distributions of the relative errors in the fraction of vacant catalytic sites $[S]$ obtained with the PINN model with the proposed post-processing and original (baseline) transformations
  }
\end{figure}

The error metrics of the PINN models with the proposed post-processing transformation \eqref{Gy-final-text} as well as with a baseline one \eqref{eq:original-post-proc-transfrm} obtained on the general testing dataset are provided in Table \ref{tab:general-spinn-metrics}.
\begin{table*}
  \centering
  \caption{Comparison of the formal accuracy metrics for SPINN models with the baseline \cite{our-prev-arxiv-preprint} and proposed post-processing transformations, obtained for the dataset with partial pressures above $\sim$ 0.32 MPa}
  \label{tab:general-spinn-metrics}
\begin{tabular}{|c|c|c|c|c|c|c|}
\hline
\multirow{2}{*}{Model} 
 & 
\multicolumn{3}{c|}{
Absolute error 
($| [S]^{\tmop{true}} - [S]^{\tmop{predicted}} |$)
}
& 
\multicolumn{3}{c|}{ 
Relative error 
($\left| \frac{[S]^{\tmop{true}} -  [S]^{\tmop{predicted}}}{[S]^{\tmop{true}}} \right|$)
}
\\
\cline{2-7}
& Mean & Median & Max & Mean & Median & Max\\
\hline
Proposed & 2.64e-05 & 2.64e-05 &  3.93e-04 &  6.55e-04 &  6.55e-04 & 5.19e-03  \\
\hline
Baseline \cite{our-prev-arxiv-preprint} & 2.31e-05 & 2.31e-05 & 1.90e-03 & 6.83e-04 & 6.83e-04 & 1.94e-02 \\
\hline
\end{tabular}
\end{table*}
In addition, Fig. \ref{fig:err_distr_gen_ds} compares the models accuracy in terms of the distribution of relative errors.
In can be seen from these data that post-processing transformation \eqref{Gy-final-text} performs similar or better than the baseline one.

\begin{figure*}
%
%
\begin{tabular*}{\textwidth}{p{0.5\linewidth}c}
\multicolumn{2}{c}{
    \begin{minipage}{0.49\linewidth}
        \includegraphics[trim={0 0 0 40},clip,width=\linewidth]{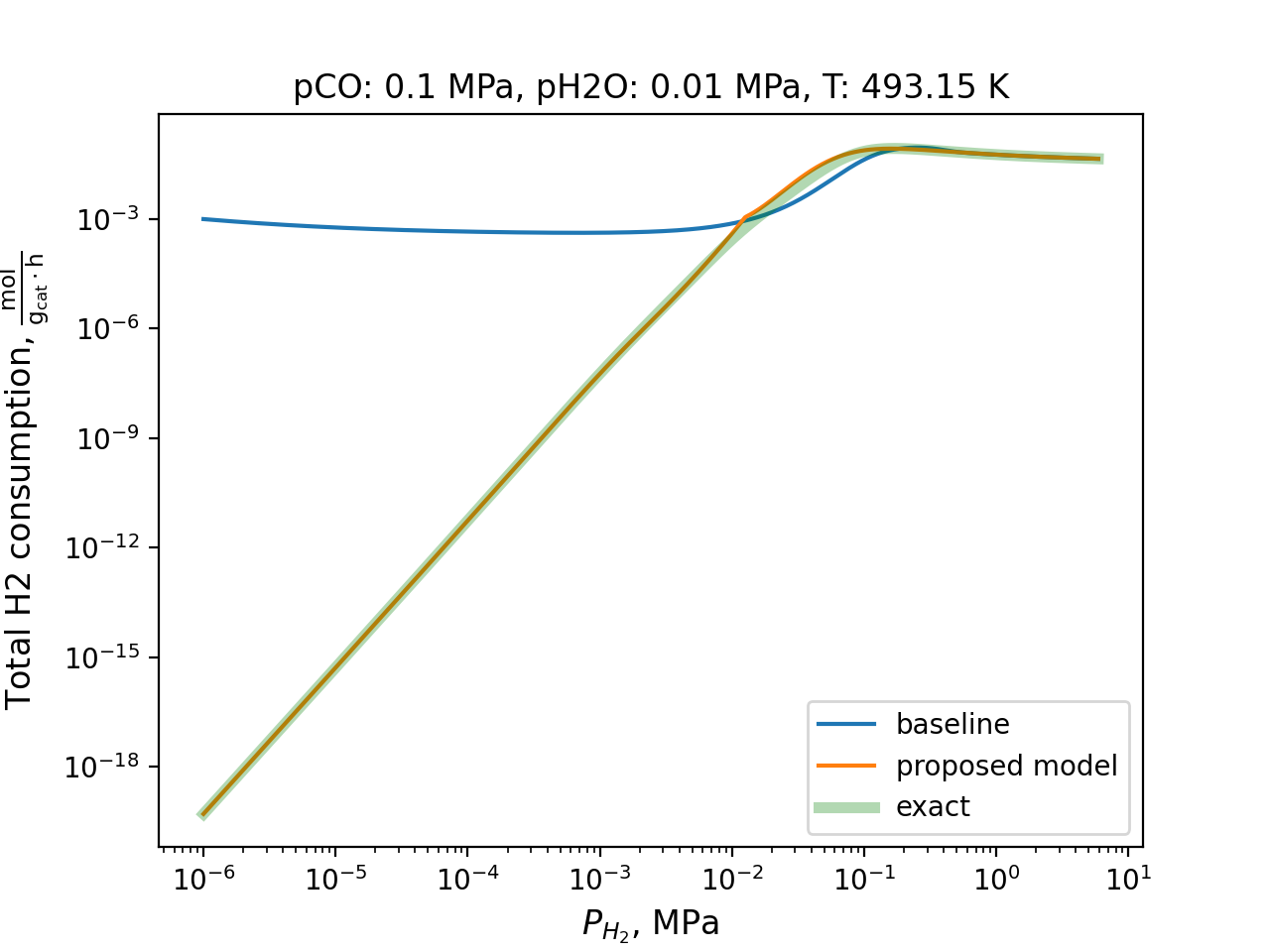} 
    \end{minipage}
    \hfill
    \begin{minipage}{0.49\linewidth}
        \includegraphics[trim={0 0 0 40},clip,width=\linewidth]{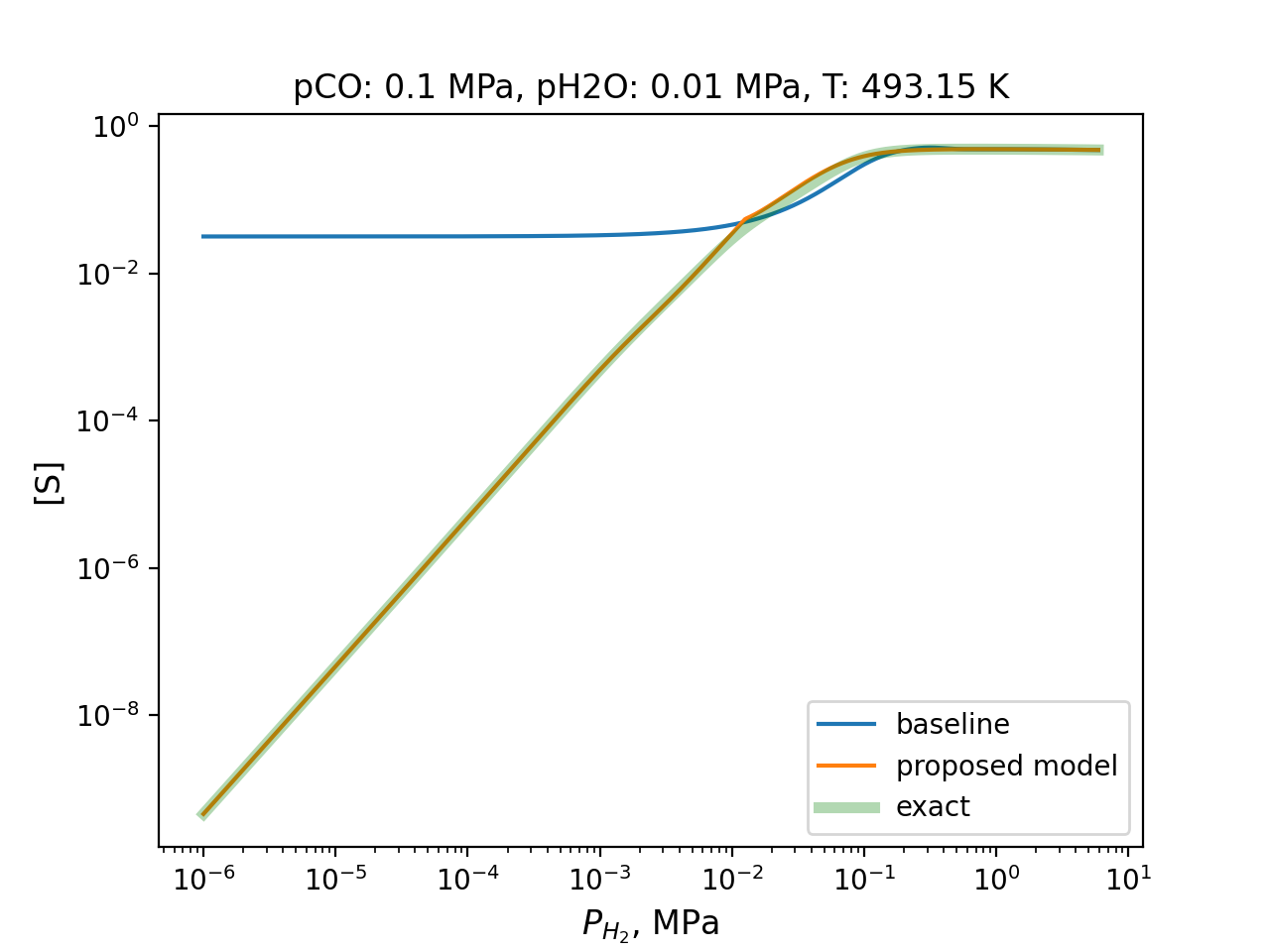} 
    \end{minipage}
    
}\\
    \centering {\textit a} & {\textit b} \\
\end{tabular*}   
  \caption{\label{fig:SPINN-performance-small-pH2-plot}Dependence of the total $H_2$ consumption rate ({\textit a}) and the fraction of vacant catalitic sites ({\textit b}) on the partial pressure of $H_2$
  (at $P_{CO} = 0.1 \; \mathrm{ MPa}$, $P_{H_2 O} = 0.01 \; \mathrm{ MPa}$, $T = 493.15 \; \mathrm{ K}$), as predicted by the ground truth solution (`exact'), PINN model with $G(y) = 10^{y}$ post-processing transformation (`baseline') \cite{our-prev-arxiv-preprint}, and the proposed PINN model which ensures correct asymptotic behaviour (`proposed model')
  }
\end{figure*}

However, the true benefits of the proposed transformation become evident when analyzed in small pressures region, not covered by the `general' testing dataset.
In Fig. \ref{fig:SPINN-performance-small-pH2-plot}, we compare the dependencies on $P_{H_2}$ obtained for the value of $[S]$ predicted by the PINN model with the proposed post-processing transformation and with the baseline one.
This figure also contains the similar dependence for the the total $H_2$ consumption rate which is obtained from the predicted $[S]$ by \eqref{R-H2-fin-tot}.
It can be seen that in contrast to the baseline approach, using the PINN model with post-processing transformation \eqref{Gy-final-text} which ensures a correct [S] asymptotics solves the issue with non-vanishing $H_2$ consumption rates in the low-pressure region.
This improvement enables the use of the PINN model in multi-stage simulations, specifically, as a component for computing the source terms in the reaction-diffusion equation.



\subsection{Initial guess for a conventional equation solver
}
\label{appendix-conv-solver-initial-guess}

Due to its iterative nature, a numerical solver applied to the reaction-diffusion equation needs an initial guess for the functions of interest to start operating.
In case of the problem at hands \eqref{wj-eq-final}, the simplest option available for the initial guess is the constant equal to the value the concentration has at the pellet boundary.
However, it appears that with such a naive guess, SciPy's {\tt solve\_bvp} with default settings converges in only ca. 30\% of all cases, as we have found by running it for all possible combinations of 50 values of $P_{CO}$ and $P_{H_2}$ equally spaced within 0 to 6 MPa and with $P_{H_2 O}$ fixed to 0.5 MPa, $T = 493.15 \mathrm{K}$.
\RemoveForNow{
{\color{red} TODO: verify if H2O pressure here is correct!}
}
The possible reasons for such a high failure rate with the naive guess are discussed in Appendix \ref{appendix-source-terms-approx}.
Specifically, we point out the shortcomings related to using a truncated Taylor series expansion for the source terms when building an implicit finite-difference scheme for the reaction-diffusion system of interest.
To mitigate these issues and ensure the solver stability, particularly, preventing it from evaluating the source terms at unphysical negative concentrations, we propose an alternative problem-specific approach to creating the initial guess.

To formulate this approach, we introduce the superscripts
$^{\tmop{old}}$ and $^{\tmop{new}}$ to denote the values of the target
quantity at the two subsequent iterations, and use
column-vector $\mathbf{w}$ to collectively represent its values at the points of some discrete grid covering the pellet.
Given that grid, we also introduce matrix $\mathbf{L}$ so that the column-vector $\mathbf{L} \mathbf{w}$ contains the values of Laplacian of the target quantity at the grid points.
The proposed method is built based on derivative-free linear approximation 
\begin{multline}
  s (w^{\tmop{new}}) = s (w^{\tmop{old}}) + \frac{s (w^{\tmop{old}}) - s
  (0)}{w^{\tmop{old}} - 0} \cdummy (w^{\tmop{new}} - w^{\tmop{old}}) 
  \EqLineBreak
  = \frac{s
  (w^{\tmop{old}})}{w^{\tmop{old}}} \cdummy w^{\tmop{new}},
  \label{sw-linear-approx-we-use-text}
\end{multline}
for the dependence of the source term $s$ on $w^{\tmop{new}}$, applied at each of the grid points.
This approximation, enables the use of an implicit numerical scheme and can be viewed either as devising a ‘local’ `slope' $\alpha = \frac{- s}{w}$ for $s(w)$ at each point of the pellet and then using it to approximate $s (w^{\tmop{new}})$, or as a result of imposing the constraint $s (w = 0) = 0$ and then building a two-point linear approximation for $s$, leading to
$\frac{s (w^{\tmop{old}}) - s (0)}{w^{\tmop{old}} - 0}$
as the value of
$\left(\frac{\partial s}{\partial w} \right)_{w^{\tmop{old}}}$
in Taylor-series expansion around $w^{new}$. 

It is worth noting that approximation (\ref{sw-linear-approx-we-use-text}) would have been unusable 
if $\lim_{w \rightarrow 0} \frac{s (w)}{w}$ was not finite.
Therefore,
when NN-based model is used to evaluate $s (w)$, such model thus must ensure correct asymptotic behaviour in small-concentrations regions
%
, so that the ratio $\frac{s (w)}{w}$ remains finite 
%
when $w \rightarrow 0$ (cf. subsection \ref{subsection-asympt-an}).
%


With (\ref{sw-linear-approx-we-use-text}), we obtain a system of linear equations
\begin{equation}
  \left( \frac{s (w^{\tmop{old}}_j)}{w^{\tmop{old}}_j} - \frac{1}{\tau}
  \right) w_j^{\tmop{new}} + (\mathbf{L} \mathbf{w}^{\tmop{new}})_j = -
  \frac{w_j^{\tmop{old}}}{\tau} \label{eq-wnew-from-wold-working-eq-text}
\end{equation}
which is used to find all components of $\mathbf{w}^{\tmop{new}}$ from $\mathbf{w}^{\tmop{old}}$ at each of the iterations (see Appendix \ref{appendix-source-terms-approx} for a detailed derivation).

As a result, the proposed initial guess creation procedure can thus be formulated as following (see also \ref{appendix-pseudocode-guess} for a pseudocode formulation):
\begin{itemize}
  \item Initialization: build the Laplacian matrix $\mathbf{L}$ for the
  selected grid, set $w_j^{\tmop{old}} = w_{\tmop{BC}}$ for all $j$
  (grid-point indices) and set $\tau_{\tmop{initial}} = \varepsilon \cdummy
  \frac{w_{\tmop{BC}}}{- s (w_{\tmop{BC}})}$, where $\varepsilon = 0.1$.
  
  \item Iterate over ‘time-steps’:
  \begin{itemize}
    \item If some components of $\mathbf{w}_j$ are negative, replace them with
    $10^{- 5}$, compute source terms, and set to zero all those components of
    the source terms $\mathbf{s} (\mathbf{w})$ which correspond to negative
    ‘concentrations’;
    
    \item Find $\mathbf{w}^{\tmop{new}}$ by solving the system of linear
    equations (\ref{eq-wnew-from-wold-working-eq-text}), with the rows of equation
    matrix which correspond to the pellet center and the outer radius of the
    pellet being replaced with the equations expressing the corresponding
    boundary conditions;
    
    \item Use non-negative components of $\mathbf{w}^{\tmop{new}}$ to compute
    the new source terms;
    
    \item If there are no negative components of $\mathbf{w}_j^{\tmop{new}}$
    AND relative (component-wise) change in $\frac{s_j}{w_j}$ (where only the
    components with non-negative $w^{\tmop{old}}$ and $w^{\tmop{new}}$ are
    considered) is less than 0.25, treat the current iteration as completed,
    increase ‘time-step’ (set $\tau \assign 2 \tau$) and proceed to the
    next iteration;
    
    Otherwise, select a smaller time-step ($\tau \assign \frac{\tau}{2}$) and
    repeat the current iteration again;
  \end{itemize}
  \item Stop if the maximum relative change in $\mathbf{w}_j$ is below 0.01,
  or if the total number of iterations exceeded 250.
\end{itemize}

Several notes are worth adding regarding the proposed algorithm.
\begin{itemize}
  \item The `time-step' $\tau$ can be viewed as a regularizing parameter, in a sense that by choosing it to be small enough, one can ensure that all the eigenvalues of the linear equations system matrix in  (\ref{eq-wnew-from-wold-working-eq-text}) are negative. In that case, it is guaranteed that all components of $\mathbf{w}^{\tmop{new}}$ are non-negative, given that the same was ensured for $\mathbf{w}^{\tmop{old}}$.  Indeed, 
  $(\mathbf{w}^{\tmop{old}})^T \mathbf{w}^{\tmop{new}} = \sum_k  (\mathbf{w}^{\tmop{old}})^T \frac{\mathbf{v}_k \mathbf{v}_k^T}{- \Lambda_k}  \mathbf{w}^{\tmop{old}} = \sum_k \frac{(\mathbf{v}_k^T  \mathbf{w}^{\tmop{old}})^2}{- \Lambda_k} \geqslant 0$ (where $\Lambda_k < 0 $
  and
  $\mathbf{v}_k$ are the eigenvalues and eigenvectors of the equation system  matrix, and $ \sum_k \frac{\mathbf{v}_k \mathbf{v}_k^T}{ \Lambda_k} $ is its inverse) is true in general, particularly, in case when just a single  component of $\mathbf{w}^{\tmop{old}}$ is kept non-zero. That being showed,  one can now represent $\mathbf{w}^{\tmop{new}}$ as a sum of ‘partial’  solutions, each corresponding to $\mathbf{w}^{\tmop{old}}$ with a single  non-zero component, thus ensuring that all components of
  $\mathbf{w}^{\tmop{new}}$ 
  are non-negative.
  
  \item The rationale behind the choice of the initial ‘time-step’ in our
  approach is the following. When all components of $\mathbf{w}^{\tmop{old}}$
  are initialized to the constant (equal to the boundary conditions
  concentration $w_{\tmop{BC}}$), we have $\mathbf{L}  \mathbf{w}^{\tmop{old}}
  \equiv 0$ in the inner points and can use ‘explicit scheme’-type
  equation (\ref{wnew-wold-explicit-1}) to \tmtextit{estimate
  }$\mathbf{w}^{\tmop{new}}$. That yields $\frac{w^{\tmop{new}} -
  w_{\tmop{BC}}}{\tau} = s (w_{\tmop{BC}})$ (notice that there is no
  dependence on $j$ because all components of $\mathbf{w}^{\tmop{old}}$ are
  the same), and hence the \tmtextit{relative} change in the concentration
  under the given $\tau$ is $\left| \frac{w^{\tmop{new}} -
  w_{\tmop{BC}}}{w_{\tmop{BC}}} \right| = \tau \left| \frac{s
  (w_{\tmop{BC}})}{w_{\tmop{BC}}} \right|$. By constraining it to be smaller
  than the chosen $\varepsilon$, we get $\tau < \varepsilon \cdummy
  \frac{w_{\tmop{BC}}}{- s (w_{\tmop{BC}})} = \tau_{\max}$, as used in the
  algorithm. The similar derivations can be done also for (\ref{eq-wnew-from-wold-working-eq-text}) by assuming that $\mathbf{L} 
  \mathbf{w}^{\tmop{new}} \approx \mathbf{L}  \mathbf{w}^{\tmop{old}} \equiv
  0$, which leads to $\left| \frac{w^{\tmop{new}} -
  w_{\tmop{BC}}}{w_{\tmop{BC}}} \right| = \frac{- \tau s
  (w_{\tmop{BC}})}{w_{\tmop{BC}} - \tau s (w_{\tmop{BC}})} < \varepsilon
  \Longrightarrow \tau < \varepsilon \frac{w_{\tmop{BC}}}{- s (w_{\tmop{BC}})
  \cdummy (1 - \varepsilon)}$, the upper bound which is \tmtextit{less}
  restrictive than the one used in the algorithm.
\end{itemize}
The proposed initial guess has been found to be quite accurate, so that on average only 1.03 iterations of a conventional solved is typically needed to reach a converged solution.
The differences between 
the guess and 
the final solution
have typically been found within 1\% (see Appendix \ref{appendix-profiles-rel-err} for a detailed comparison).

To showcase the role of the initial guess quality on the convergence of the reaction-diffusion equation numerical solver,  we considered 2500 sets of boundary condition values created by combining 50 values of $P_{CO}$ and $P_{H_2}$, each equally spaced within 0 to 6 MPa range, with the fixed values of $P_{H_2 O} = 0.5 \mathrm{MPa}$ and $T = 493.15 \mathrm{K}$.
\RemoveForNow{
{\color{red} TODO: verify whether the ranges/values are correct here }
}
For each of these boundary conditions, the proposed initial guess generator in conjunction with SciPy's {\tt solve\_bvp} with default settings was used to find a `ground truth' solutions $w^{\tmop{exact}}_j(x)$ of the reaction-diffusion equations.
The process converged in all 100\% cases.

We then repeated a similar process of running SciPy's {\tt solve\_bvp} on the same sets of boundary conditions, but this time used the following expression as an initial guess
\begin{equation}
w_j (x) = w_j^{\tmop{BC}} + (w_j^{\tmop{exact}} (x) - w_j^{\tmop{BC}}) \cdummy \gamma
\end{equation}
where we introduced $\gamma$ which can be interpreted as the `quality' of an initial guess, in a sense that $\gamma = 0$ corresponds to the naive guess with all concentrations set equal to their values on the pellet boundary, while $\gamma = 1$ provides the guess equal to `exact' solution.
The analysis of the solver convergence statistics was then performed.
Success rate of the solver convergence was judged by two metrics: the fraction of all sets of boundary conditions at which the solver converged, and the fraction of all sets of boundary conditions when the obtained solution did not contain negative (unphysical) values of the concentrations.
Additionally, we analyzed the total running time of the solver as well as the number of iterations it needed to reach to the final solution.
\begin{figure*}[h]
{
\begin{tabular*}{\textwidth}{
p{0.5\linewidth}c
}
\multicolumn{2}{c}{
\includegraphics[width=0.9\textwidth]{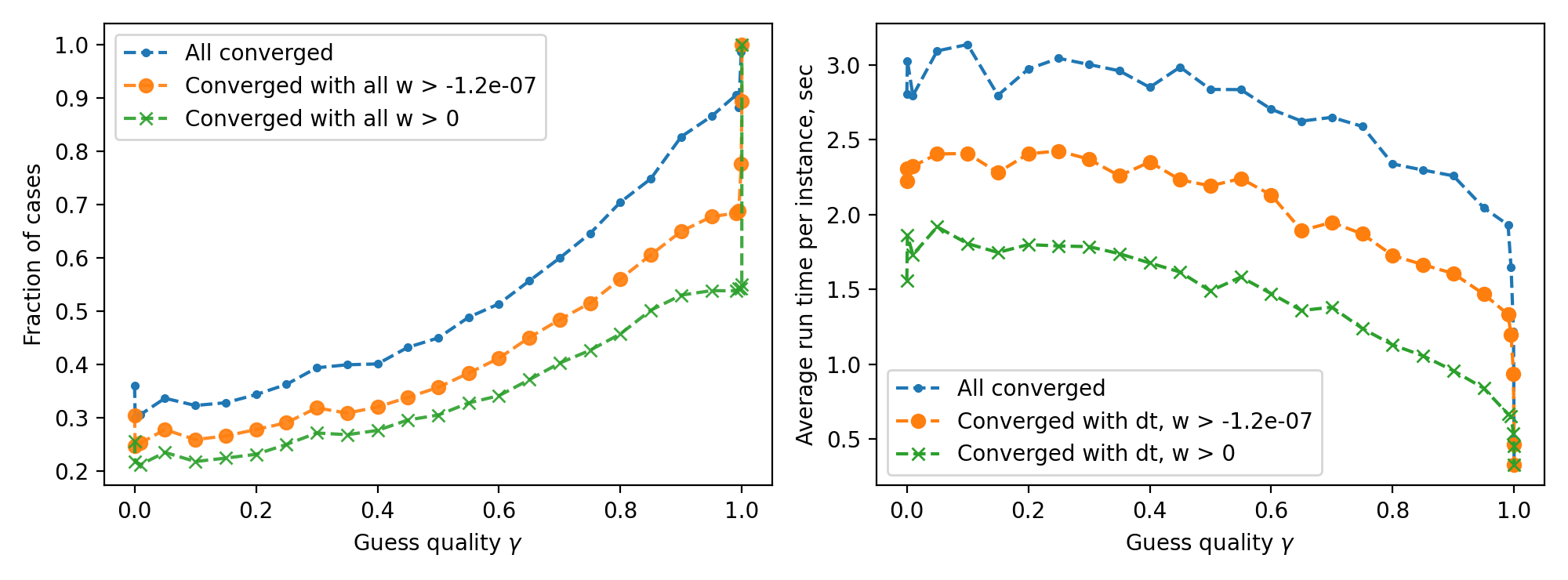}
}\\
    \centering {\textit a} & {\textit b} \\
\end{tabular*}   
 }  
  \caption{
  \label{fig-guess-quality-effect1}
  Effect of the initial guess quality $\gamma$ on the convergence of the finite-difference ODE solver applied to the studied reaction-diffusion system with different values of boundary conditions: the fraction of cases in which the solver converges ({ \textit a}) and the average wall-time spent to solve one instance, not including the time needed to generate initial guess ({\textit b}). The cases in which no unphysical values are found in the concentration profiles are also shown for convenience: `$w>0$' denotes the cases in which the found profiles contain no negative values, while `$w > \mathrm{-1.2e-07}$' corresponds to the similar condition released so that all negative values within the machine epsilon round-off threshold of $2^{-23} \approx 1.2\cdot 10^{-7}$ are considered as still physically meaningful.
  }
\end{figure*}
As shown in Fig. \ref{fig-guess-quality-effect1}, the quality of the initial guess is crucial to ensure the convergence of the solver, highlighting the practical usefulness of the proposed initialization scheme.
In particular, it can be seen that in order to bring the ratio of successfully converged cases close to 100\%, initial guess should differ from exact solution within a few percents.
It is thus no surprise that the initial guesses $w^{guess}_j(r)$ found by the method presented in subsection \ref{appendix-conv-solver-initial-guess} are rather close to the ground truth solution $w^{final}_j(r)$, ($j = \mathrm{CO}, \; \mathrm{H_2}$), found by the numerical solver.
\RemoveForNow{
\tmcolor{blue}{* ???2d-figure: time stats}

\tmcolor{blue}{\{+?provide statistics on its accuracy as compared to ‘full’ solver, and the average number of iterations needed\}}
}
%
%
%
%

\section{Conclusions}

Design of a neural network trained in a physics-informed manner has been adapted to be usable for accelerating the solution of a multi-stage simulation problem related to Fischer-Tropsch gas-to-liquid chemical synthesis process.
We have demonstrated that when the catalyst pellet
as the key component of the chemical reactor running the synthesis is described with the system of reaction-diffusion equations, special care must be taken when replacing their exact source terms with numerically more efficient approximations, mimicking exact microkinetics model.
Conventional feed-forward architecture of a neural network used as a core component for such an approximation has been found to be inapplicable, as it failed to reproduce correct asymptotic of the reaction rates under small reactants pressures conditions.
To tackle this obstacle, we proposed a post-processing transformation, effectively acting as  an additional problem-specific output layer of a neural network.
We have found that with a properly designed initialization scheme, the proposed PINN architecture allows  efficiently solving the system of reaction-diffusion equations for wide range of  thermodynamics conditions on the boundary of the pellet. 
The proposed combination of the PINN architecture and numerical solver initialization scheme can be further applied for ground-up modeling of FTS reactors, allowing for theory-based computation of the properties of chemical interest.



\putReferencesList

\appendix

\section{
Corrections extrapolating the overall reaction rates to an infinite $N_{\max}$
} \label{appendix-to-inf-corrs}

  In order to compensate for the error appearing in (\ref{R-CO-fin-tot}) and (\ref{R-H2-fin-tot}) due to the finite upper bound $N_{\max}$ of summation, we first notice that each of the individual production rates given by (\ref{eq:R_CnH2np2})--(\ref{eq:R_C2H4}) is proportional to $n \cdummy \prod_{j = 1}^n \alpha_j$ (in case of paraffins) or $n \cdummy e^{c n} \cdummy \prod_{j = 1}^n \alpha_j$ (in case of olefins). Specifically,
  \begin{multline}   s_{\tmop{par}} = \frac{\sum_{n = 1}^{\infty} (n \cdot R_{\mathrm{C}_n   \mathrm{H}_{2 n + 2}}) - \sum_{n = 1}^{N_0} (n \cdot R_{\mathrm{C}_n   \mathrm{H}_{2 n + 2}})}{\kappa_{\tmop{par}, \tmop{long}} \cdummy   [\mathrm{S}]^2} 
  \EqLineBreak
  = \sum_{n = N_0 + 1}^{\infty} \left( n \cdot \prod_{j = 1}^n   \alpha_j \right) \label{s-par-corr-def} \end{multline}
  and 
  \begin{multline}   s_{\tmop{ole}} = \frac{\sum_{n = 2}^{\infty} (n \cdot R_{\mathrm{C}_n   \mathrm{H}_{2 n}}) - \sum_{n = 2}^{N_0} (n \cdot R_{\mathrm{C}_n   \mathrm{H}_{2 n}})}{\kappa_{\tmop{ole}, \tmop{long}} \cdummy [\mathrm{S}]} 
  \EqLineBreak
  =   \sum_{n = N_0 + 1}^{\infty} \left( n \cdot e^{c \cdot n} \cdot \prod_{j =   1}^n \alpha_j \right) \label{s-ole-corr-def}
  \end{multline}
can be introduced as the corrections related to truncations of the sums. Summation bound $N_0$ is introduced here as the threshold defined so that for $j > N_0$ the chain growth probabilities $\alpha_j$, defined by (\ref{alpha-n3-def}), are essentially independent on $j$ as a consequence of $e^{c \cdummy j}$ becoming vanishingly small in (\ref{alpha-n3-def}). For $j > N_0$ we can thus put $\alpha_j \approx \alpha_{N_0 + 1}$.   
It is further assumed that $N_{\max} \geqslant N_0$, i.e., $N_{\max}$ used in (\ref{R-CO-fin-tot}) and (\ref{R-H2-fin-tot}) exceeds the threshold $N_0$.
As (\ref{s-par-corr-def}) can be considered as a partial case of (\ref{s-ole-corr-def}) with $c = 0$, it is sufficient to analyze (\ref{s-ole-corr-def}) only.
By replacing $\alpha_j$ with $\alpha_{N_0 + 1}$ for $j \geqslant N_0 + 1$, we get for $n \geqslant N_0 + 1$ 
\begin{multline}
 e^{c n}  \prod_{j = 1}^n \alpha_j = \prod_{j = 1}^{N_0} \alpha_j \cdot e^{c    n}  \prod_{j = N_0 + 1}^n \alpha_j 
 \approx  e^{c n} (\alpha_{N_0 + 1})^{n - N_0}  \prod_{j = 1}^{N_0} \alpha_j
 \EqLineBreak
 = \frac{\prod_{j = 1}^{N_0}    \alpha_j}{\alpha_{N_0 + 1} \, ^{N_0}} \cdot (\alpha_{N_0 + 1} e^c)^n 
\end{multline}
It is convenient to introduce $\tilde{\alpha}_{N_0 + 1} = \alpha_{N_0 + 1} e^c$ as a shorthand notation.
We thus have to compute 
\begin{multline}
  s_{\tmop{ole}} \approx \sum_{n = N_0 + 1}^{\infty} \left( n \cdot   \frac{\prod_{j = 1}^{N_0} \alpha_j}{\alpha_{N_0 + 1} \,^{N_0}} \cdot   \tilde{\alpha}_{N_0 + 1} \,^n \right) 
  \EqLineBreak
  =   \frac{\prod_{j = 1}^{N_0}   \alpha_j}{\alpha_{N_0 + 1} \,^{N_0}} \cdummy \sum_{n = N_0 + 1}^{\infty} n   \cdot \tilde{\alpha}_{N_0 + 1} \,^n = s_{\tmop{ole}}^{\tmop{approx}}   \label{s-ole-approx-def} 
\end{multline}
The sum \ $\sum_{n = N_0 + 1}^{\infty} n \cdot \tilde{\alpha}_{N_0 + 1} \,^n = \tilde{\alpha}_{N_0 + 1} \,^{N_0} \cdummy \sum_{k = 1}^{\infty} (N_0 + k) \cdot (\tilde{\alpha}_{N_0 + 1})^k$ can be computed by first using $\sum_{k = 1}^{\infty} \tilde{\alpha}_{N_0 + 1} \,^k = \frac{\tilde{\alpha}_{N_0 + 1}}{1 - \tilde{\alpha}_{N_0 + 1}}$ as the sum of geometric series, and then by taking derivative of this result with respect to $\tilde{\alpha}_{N_0 + 1}$. 
That yields
\[ \sum_{n = 1}^{\infty} n \cdot \tilde{\alpha}_{N_0 + 1} \,^n =    \frac{\tilde{\alpha}_{N_0 + 1}}{(1 - \tilde{\alpha}_{N_0 + 1})^2} . 
\] 
Combined together, these results imply 
\begin{multline}
\sum_{n = N_0 + 1}^{\infty} n \cdot (\tilde{\alpha}_{N_0 + 1})^n = 
\EqLineBreak
\tilde{\alpha}_{N_0 + 1} \,^{N_0} \cdummy \left( N_0  \left( \frac{1}{1 -   \tilde{\alpha}_{N_0 + 1}} - 1 \right) 
 +  \frac{\tilde{\alpha}_{N_0 + 1}}{(1 -   \tilde{\alpha}_{N_0 + 1})^2} \right) 
\\
 =  \tilde{\alpha}_{N_0 + 1} \,^{N_0} \cdummy L (\tilde{\alpha}_{N_0 +   1}, N_0)
\label{sum-n-alphatilde-n} 
\end{multline}
%
where we introduced a function (\ref{Lfunc-def}) for the sake of convenience.
With (\ref{sum-n-alphatilde-n}) we have from (\ref{s-ole-approx-def})
\begin{multline}
s_{\tmop{ole}}^{\tmop{approx}} = \tilde{\alpha}_{N_0 + 1} \,^{N_0} \cdummy   \frac{\prod_{j = 1}^{N_0} \alpha_j}{\alpha_{N_0 + 1} \,^{N_0}} \cdummy L   (e^c \alpha_{N_0 + 1}, N_0) 
\EqLineBreak
= e^{c N_0} \cdummy L (e^c \alpha_{N_0 + 1},   N_0) \cdummy \prod_{j = 1}^{N_0} \alpha_j 
\label{s-ole-approx-res1} 
\end{multline}
and, by substituting $c = 0$, the similar result for paraffins becomes
\begin{multline}
s_{\tmop{par}}^{\tmop{approx}} = L (\alpha_{N_0 + 1}, N_0) \cdummy \prod_{j   = 1}^{N_0} \alpha_j 
\EqLineBreak
= L (\alpha_{N_0 + 1}, N_0) \cdummy \frac{R_{C_{N_0}   H_{2 N_0 + 2}}}{\kappa_{\tmop{par}, \tmop{long}} \cdummy [\mathrm{S}]^2} ,
\label{s-par-res2}
\end{multline}
where we used (\ref{eq:R_CnH2np2}). Similarly to that, (\ref{eq:R_C2H2n}) can be used to represent (\ref{s-ole-approx-res1}) as
\begin{equation}
s_{\tmop{ole}}^{\tmop{approx}} = L (e^c \alpha_{N_0 + 1}, N_0) \cdummy   \frac{R_{C_{N_0} H_{2 N_0}}}{\kappa_{\tmop{ole}, \tmop{long}} \cdummy   [\mathrm{S}]} \label{s-ole-approx-res2} 
\end{equation}
To estimate a lower bound for $N_0$, we should require that $e^{cN_0} \ll 1$. 
According to Eq.(6) of {\cite{Todic-CO-insertion-2014}}, $c = - \hspace{0.27em} \frac{\Delta E}{RT}$, with Table 2 of the same paper suggesting $\Delta E = 1.1 \frac{\tmop{kJ}}{\tmop{mol}}$, so that for 513 K as the upper boundary of the selected temperature range, we have $RT / \Delta E \approx 4$.
Then, if we require that $- cN_0 = - \ln 10^{- 8} \approx 18$ (with the `small value' of $10^{- 8}$ being related to the machine epsilon round-off threshold of $2^{-23} \approx 1.2\cdot 10^{-7}$ in case of 32-bit single precision representation of floating-point numbers), we conclude that any $N_0 > 4 \cdot 18 = 72$ should be a reasonable choice. We thus simply set $N_0$ equal to $N_{\max} = 100$.

When combined with (\ref{s-par-corr-def}) and (\ref{s-ole-corr-def}), results (\ref{s-par-res2}) and (\ref{s-ole-approx-res2}) now lead to the approximations
\begin{multline}
\sum_{n = 1}^{\infty} (n \cdot R_{\mathrm{C}_n \mathrm{H}_{2 n + 2}})   \approx \sum_{n = 1}^{N_{\max}} (n \cdot R_{\mathrm{C}_n \mathrm{H}_{2 n +   2}}) 
\EqLineBreak
+ L (\alpha_{N_{\max} + 1}, N_{\max}) \cdummy R_{C_{N_{\max}} H_{2   N_{\max} + 2}} 
\label{inf-sum-n-Rpar}
\end{multline}
for paraffins and
\begin{multline}
\sum_{n = 2}^{\infty} (n \cdot R_{\mathrm{C}_n \mathrm{H}_{2 n}}) \approx   \sum_{n = 2}^{N_{\max}} (n \cdot R_{\mathrm{C}_n \mathrm{H}_{2 n}}) 
\EqLineBreak
+ L (e^c   \alpha_{N_{\max} + 1}, N_{\max}) \cdummy R_{C_{N_{\max}} H_{2 N_{\max}}}   \label{inf-sum-n-Role} 
\end{multline}
for olefins.
Furthermore, as discussed in {\cite{our-prev-arxiv-preprint}}, 
$$
\sum_{n = N_0 + 1}^{\infty} \prod_{j = 1}^n \alpha_j \approx \prod_{j =    1}^{N_0} \alpha_j \cdot \hspace{0.27em} \frac{\alpha_{N_0 + 1}}{1 -    \alpha_{N_0 + 1}} 
$$
implying that 
\begin{equation}
\sum_{n = 1}^{\infty} R_{\mathrm{C}_n \mathrm{H}_{2 n + 2}} - \sum_{n =   1}^{N_{\max}} R_{\mathrm{C}_n \mathrm{H}_{2 n + 2}} \approx   R_{\mathrm{C}_{N_{\max}} \mathrm{H}_{2 N_{\max} + 2}} \cdot \hspace{0.27em}   \frac{\alpha_{N_{\max} + 1}}{1 - \alpha_{N_{\max} + 1}} .   \label{inf-sum-Rpar} 
\end{equation}
Results (\ref{inf-sum-n-Rpar}), (\ref{inf-sum-n-Role}) and (\ref{inf-sum-Rpar}) thus allow extrapolating $R_{CO}$ and $R_{H_2}$, defined in (\ref{R-CO-fin-tot}) and (\ref{R-H2-fin-tot}), to $N_{\max} = \infty$.



%
%
%
%
%
%

\section{Heuristics used to construct a post-processing transformation for SPINN}
\label{post-processing-heuristics-appendix}

We now turn to discussing some heuristics which are helpful for building the function $G$ capable of ensuring that $[S]$ has a correct asymptotic behaviour in the small pressures region and, therefore, the required extrapolation abilities.
By considering (\ref{eq-S-via-J}) as a starting point, the simplest form of the post-processing transformation would be
$$
G^{\tmop{simple}} (y) = \frac{1}{c_0 + c_S \cdummy \alpha_1 \cdummy (1 +
   y)} 
$$
so that with 
$y = \tmop{SPINN} (\bar{X})$ we get $[S] = \left( c_0 + c_S
\cdummy \alpha_1 \cdummy (1 + \tmop{SPINN} (\bar{X}))\right)^{-1}$. 
However, such form can require SPINN to output values much larger than unity, which is not desirable, as it would deviate the output from a standard normal distribution significantly. 
Another option could be using 
$\frac{\alpha_{\infty}}{1 -
\alpha_{\infty}}$,
in which $0 < \alpha_{\infty} < 1$ is predicted by SPINN, to approximate $J$ in (\ref{eq-S-via-J}) instead of $1 + y$, but such a choice would neglect the difference between $\kappa_{\tmop{ole}, \tmop{short}}$ (as present in $\alpha_2$) and $\kappa_{\tmop{ole}, \tmop{long}}$ (present in $\alpha_n$ with $n \geqslant 3$).

We thus constructed the following two-staged alternative to create transformation $G$.
First, we use (\ref{alpha1-def-kapp}) for $\alpha_1$ and set $\alpha_{\infty} = \frac{1}{1 + \varepsilon}$, where 
$\varepsilon = \frac{\kappa_{\tmop{par}, \tmop{long}}}{\kappa_{\tmop{growth}}}$ 
as before (cf. \eqref{eq-epsilon-kappa-ratio-def}). 
As $\alpha_{\infty} > \alpha_n$ for all $n > 2$, 
$\alpha_{\infty} + \alpha_{\infty}^2 + \alpha_{\infty}^3 + \cdots
= \frac{\alpha_{\infty}}{1 - \alpha_{\infty}} = \frac{1}{\varepsilon} =
\frac{\kappa_{\tmop{growth}}}{\kappa_{\tmop{par}, \tmop{long}}}$ 
can be used as ‘an order of magnitude estimate’ for $J - 1$, or, more strictly, as an upper bound for 
$\alpha_2 + \alpha_2 \alpha_3 + \alpha_2 \alpha_3 \alpha_4 +
\ldots$. 
This further implies that 
$0 \leqslant \frac{\alpha_2 + \alpha_2\alpha_3 + \alpha_2 \alpha_3 \alpha_4 + \ldots}{\alpha_{\infty} + \alpha_{\infty}^2 + \alpha_{\infty}^3 + \ldots} = \frac{J - 1}{\frac{\alpha_{\infty}}{1 - \alpha_{\infty}}} \leqslant 1$, 
so that the SPINN output 
$y = \tmop{SPINN} \left( \text{\tmtextit{\={X}}} \right)$
contained in 0 to 1 range can be scaled as 
$\frac{\alpha_{\infty}}{1 - \alpha_{\infty}} \cdummy y$ 
to approximate $J - 1$. As the difference between
$\kappa_{\tmop{ole}, \tmop{short}}$ 
(as present in $\alpha_2$) and
$\kappa_{\tmop{ole}, \tmop{long}}$ 
(present in $\alpha_n$ with $n \geqslant 3$) is still neglected is these considerations, the quantity
\begin{multline}
  \sigma (y) = \frac{1}{c_0 + c_S \cdummy \alpha_1 \cdummy \left( 1 +
  \frac{\alpha_{\infty}}{1 - \alpha_{\infty}} \cdummy y \right)} 
  \EqLineBreak
  =
  \frac{\kappa_{\tmop{par}, \tmop{long}}}{c_0 \cdummy \kappa_{\tmop{par},
  \tmop{long}} + c_S \cdummy \alpha_1 \cdummy (\kappa_{\tmop{par},
  \tmop{long}} + \kappa_{\tmop{growth}} \cdummy y)} \label{sigma-y}
\end{multline}
introduced at this stage is only treated as an auxiliary intermediate quantity, or just as an ‘initial estimate’ for $[S]$.

Indeed, this $\sigma$ alone is not good enough as the value of $[S]$ at very small $\varepsilon$, because assuming that SPINN output $y$, 
$0 \leqslant y \leqslant 1$, goes on a plateau at small reactants pressures, we'd have $\sim \frac{1}{\varepsilon^4} \cdummy \left( 1 + \frac{1}{\varepsilon} \cdummy y \right)$ 
asymptotics in the denominator of (\ref{sigma-y}) at small $\varepsilon$, given that 
$\frac{\alpha_{\infty}}{1 - \alpha_{\infty}} = \frac{1}{\varepsilon}$ and $c_S \sim \frac{1}{P_{H_2}^2} \sim
\frac{1}{\varepsilon^4}$. 
That would imply that $\sigma$ follows 
$\sigma \sim \varepsilon^5$ 
asymptotics, which is thus defined not only by that of $c_S$ and is in contrast to \eqref{eq-S-PH2-sq}.
In other words, $y$ can not simply become constant at small input pressures
in case if $G$ is expected to work in the ‘extrapolation’ regime. 
This drawback comes from the fact that the ‘large’ 
$\frac{\alpha_{\infty}}{1 - \alpha_{\infty}} = \frac{1}{\varepsilon}$ 
%
%
is not ‘damped’ by any other factor, which in the true $J$ comes from a number of initial $\alpha_j$'s (cf. \eqref{eq-J-asympt-factored-out-1} and \eqref{eq-alphaj-with-large-A}).

The same effect of ‘damping’ can, however, be introduced empirically by using the above-introduced $\sigma$ at one stage of the two-stage procedure.

To this end, we introduce an ‘effective value’ for $\alpha_2$ as
\begin{equation}
  \tilde{\alpha}_2 = (\alpha_2)_{[S] = \sigma} =
  \frac{\kappa_{\tmop{growth}}}{\kappa_{\tmop{growth}} + \kappa_{\tmop{par},
  \tmop{long}} + e^{- 2 c} \cdummy \frac{\kappa_{\tmop{ole},
  \tmop{short}}}{\sigma}}
\end{equation}

and then use it to get
\begin{equation}
  [S] = G (y) = \frac{1}{c_0 + c_S \cdummy \alpha_1 \cdummy \left( 1 +
  \tilde{\alpha}_2 + \tilde{\alpha}_2 \cdummy \frac{\alpha_{\infty}}{1 -
  \alpha_{\infty}} \cdummy y \right)} \label{Gy-final}
\end{equation}
When $\sigma$ is small, 
$\tilde{\alpha}_2 \approx \sigma \frac{\kappa_{\tmop{growth}}}{e^{- 2 c} \cdummy \kappa_{\tmop{ole}, \tmop{short}}} \sim \varepsilon^5$, 
which is small enough to ‘damp’
$\frac{\alpha_{\infty}}{1 - \alpha_{\infty}} \cdummy y \sim \frac{1}{\varepsilon}$, ensuring that $1 + \tilde{\alpha}_2 + \tilde{\alpha}_2 \cdummy \frac{\alpha_{\infty}}{1 - \alpha_{\infty}} \cdummy y \approx 1$ 
for small $\varepsilon$ and that the dependence of $[S]$ on $\varepsilon$ becomes
determined by $c_S \sim {P_{H_2}}^{-2}$ and thus in fact independent of $y$, as required.
When $\varepsilon$ is not as small (say, at the boundary of the interpolation region of SPINN inputs), $\sigma$ can be expected to be a good guess for $[S]$. 
This is because 
$1 + \frac{\alpha_{\infty}}{1 - \alpha_{\infty}} \cdummy y = 1 + \frac{y}{\varepsilon} \approx 1$ 
is the denominator of (\ref{sigma-y}) as soon as $\varepsilon$ becomes larger than a plateau on which SPINN output $y$ rests at small input pressures, and, as a result, $\sigma$ follows the correct $\sim \varepsilon^4$ (as determined by
$c\tmrsub{S}{\sim}{\frac{1}{P\tmrsub{H\tmrsub{2}}\tmrsup{2}}}{\sim}{\frac{1}{{\varepsilon}\tmrsup{4}}})$
asymptotics, as opposed to $\sigma \sim \varepsilon^5$ which took place at smaller $\varepsilon$. 
In turn, when $\sigma \approx [S]$, $\tilde{\alpha}_2 \approx \alpha_2$ can be anticipated as well, so that $\frac{\alpha_{\infty}}{1 - \alpha_{\infty}}$ now needs to approximate only $\alpha_3 + \alpha_3 \alpha_4 + \cdots$, instead of 
$\alpha_2 + \alpha_2 \alpha_3 + \alpha_2 \alpha_3 \alpha_4 + \ldots$ 
as it was in (\ref{sigma-y}).
This way, (\ref{Gy-final}) can properly account for the difference between $\kappa_{\tmop{ole}, \tmop{short}}$ and $\kappa_{\tmop{ole}, \tmop{long}}$, or $\alpha_2$ and $\alpha_n$ with $n \geqslant 3$ as defined by (\ref{alpha2-def}) and (\ref{alpha-n3-def}) respectively.

From the numerical implementation perspective, expression (\ref{Gy-final}) can
be more conveniently rewritten in an equivalent form as
%

    \begin{equation}
      [S] = \frac{\kappa_{\tmop{par}, \tmop{long}}}{c_0 \cdummy
      \kappa_{\tmop{par}, \tmop{long}} + c_S \cdummy \alpha_1 \cdummy
      (\kappa_{\tmop{par}, \tmop{long}} + \tilde{\alpha}_2 \cdummy
      (\kappa_{\tmop{par}, \tmop{long}} + \kappa_{\tmop{growth}} \cdummy y))} .
    \label{G-y-def-impl-form}
    \end{equation}

\section{Pseudo-code reresentation of the initial-guess creation procedure} \label{appendix-pseudocode-guess}

The algorithm is outlined as 
Algorithm \ref{alg:init} below.

\

\

\begin{algorithm}
\caption{Procedure for initial-guess creation}\label{alg:init}
\begin{algorithmic}
\State {Initialization: build the Laplacian matrix $\mathbf{L}$ for the  selected grid. $w_j^{\tmop{old}} \gets w_{\tmop{BC}}$ $\forall  j$ and $\tau_{\tmop{initial}} \gets \varepsilon \cdummy
  \frac{w_{\tmop{BC}}}{- s (w_{\tmop{BC}})}$, where $\varepsilon = 0.1$.}
\State $Counter \gets 0$
\State $Tol_1 \gets 0.25$, $Tol_2 \gets 0.01$, $N_{max} \gets 250$
\While{$Counter\le N_{max}$}
    \State $Temp = []$
    \For{$j$ in $len(\mathbf{w})$}
    \If{$\mathbf{w}_j < 0$}
        \State$\mathbf{w}_j \gets 10^{- 5}$
        \State $Temp.insert(j)$
    \EndIf
    \EndFor
    \State Compute source terms $\mathbf{s} (\mathbf{w})$
    \For{$i$ in $Temp$}
        \State$\mathbf{s} (\mathbf{w})_i \gets 0$
    \EndFor
    \State Find $\mathbf{w}^{\tmop{new}}$ by solving the system of linear
    equations (\ref{eq-wnew-from-wold-working-eq}). Replace rows of equation
    matrix which correspond to the pellet center and the outer radius of the
    pellet with the equations expressing the corresponding
    boundary conditions.
    \State Use non-negative components of $\mathbf{w}^{\tmop{new}}$ to compute
    the new source terms.
    \If{($\mathbf{w}_j^{\tmop{new}} >0 ~ \forall j$) AND ($(\frac{s_j}{w_j})^{\tmop{new}} - (\frac{s_j}{w_j})^{\tmop{old}} < Tol_1 )$}
    \State Increase time step $\tau \gets 2 \tau$ and proceed to next iteration
    \Else{ Reduce the timestep $\tau \gets \frac{\tau}{2}$ and repeat the current iteration.}
    \EndIf
    \If{$\exists j $ such that $ \delta \mathbf{w}_j \le Tol_2$}
    \State BREAK
    \EndIf
\State $Counter++$
\EndWhile
\end{algorithmic}
\end{algorithm}


\section{Application example}

The developed multi-stage numerical scheme accelerated by PINN enabled us to compute several quantities of chemical interest as a function of non-dimensionalized reactants pressures 
$  \bar{X}_{CO} =  p_{\mathrm{CO}} / p_{\mathrm{CO}}^{\max}  $
and 
$\bar{X}_{H_2} = p_{\mathrm{H}_2} / p_{\mathrm{H}_2}^{\max}  $ 
at fixed temperature of 493.15 K and $\mathrm{H_2 O}$ pressure of 0.5 MPa.
Technically, for each of the 2500 considered combinations of $\mathrm{H_2}$ and CO pressures, the concentration profiles were obtained by solving the reaction-diffusion equation \eqref{wj-eq-final}, and the resulting hydrocarbons formation rates were then analyzed.
We considered such quantities as
total consumption of CO by the pellet
\begin{equation}
    R^{tot}_{CO} = \int_0^{R_p} { R_{CO}(
    P_{\tmop{CO}} (r), P_{H_2} (r), P_{H_2 O} (r), T
    ) 4 \pi r^2 d r } 
    ,
    \label{eq-R-tot-pellet-CO}
\end{equation}
$\mathrm{C}_{5 +}$ fraction
\begin{equation}
    g_{C_{5 +}} = \frac{\sum_{n = 5}^{N_{\max}} R^{tot}_{C_n H_{2 n + 2}} + \sum_{n = 5}^{N_{\max}} R^{tot}_{C_n H_{2 n}}}{\sum_{n = 1}^{N_{\max}} R^{tot}_{C_n H_{2 n + 2}} +\sum_{n = 2}^{N_{\max}} R^{tot}_{C_n H_{2 n}}}    
    ,
    \label{c5-plus-fraction-def}
\end{equation}
and the pellet effectiveness factor for CO consumption
\begin{equation}
    \eta_{\tmop{CO}} = \frac{\int_0^{R_p} R_{\tmop{CO}} (P_{\tmop{CO}} (r), P_{H_2} (r), P_{H_2 O} (r), T) \cdummy 4 \pi r^2 d r}{\frac{4}{3} \pi R_p^3 \cdummy R_{\tmop{CO}} (P_{\tmop{CO}} (R_p), P_{H_2} (R_p), P_{H_2 O} (R_p), T )}
    .
    \label{eta-CO-def}
\end{equation}
\RemoveForNow{
{\color{red}verify whether we have exatly the same expressions in the code}
}
To this end, all the reaction rates were computed according to \eqref{eq:R_CnH2np2},  \eqref{eq:R_CH4},  \eqref{eq:R_C2H2n},  \eqref{eq:R_C2H4}, \eqref{R-CO-fin-tot}, by substituting the substances partial pressures from \eqref{cj-to-Pj-TDC} with the concentrations found at the given radial coordinate.
For $R^{tot}_{C_n H_{2 n + 2}}$ and $R^{tot}_{C_n H_{2 n}}$, denoting the total production of paraffins and 1-olefins respectively, expressions similar to \eqref{eq-R-tot-pellet-CO} were used.
Numerical integration in \eqref{eq-R-tot-pellet-CO}, \eqref{c5-plus-fraction-def}, \eqref{eta-CO-def} were performed using the trapezoid method with 100 equally-spaced radial points.

The above introduced quantities reflect the chemical performance characteristics of the catalyst pellets, which are practically important from the perspective of FTS reactors modelling.
For example, the total consumption of CO is related to average volumetric consumption rate of reactants within the reactor volume, which, in turn, would be required if their spatial distributions of substances within the reactor had to be found.
The $\mathrm{C}_{5+}$ fraction, on the other hand, evaluates the distribution of FTS products by prioritizing the paraffins and 1-olefins with the number of carbon atoms not less than 5, that is liquids and waxes.
Finally, the pellet effectiveness factor quantifies the impact of the diffusion limitations on the catalyst pellet performance.
It is a dimensionless property, limited to 0 to 1 range, which compares the actual consumption of a reactant within the pellet to the theoretical rate that would occur if the entire pellet interior were exposed to the same conditions as its surface.
Its highest possible value of 1 would correspond to the ideal situation with no diffusion limitations, when the reactant concentration profile is uniform throughout the pellet radius, so that the catalyst at each point of the pellet is equally active.
In contrast to that, $\eta_{CO}$ can be lowered to zero in case the reactant is consumed near the pellet surface, before it can diffuse to its interior, leaving most part of the pellet volume unutilized.

\begin{figure}[h]
\centering 
{
    \halfpagefgraphics[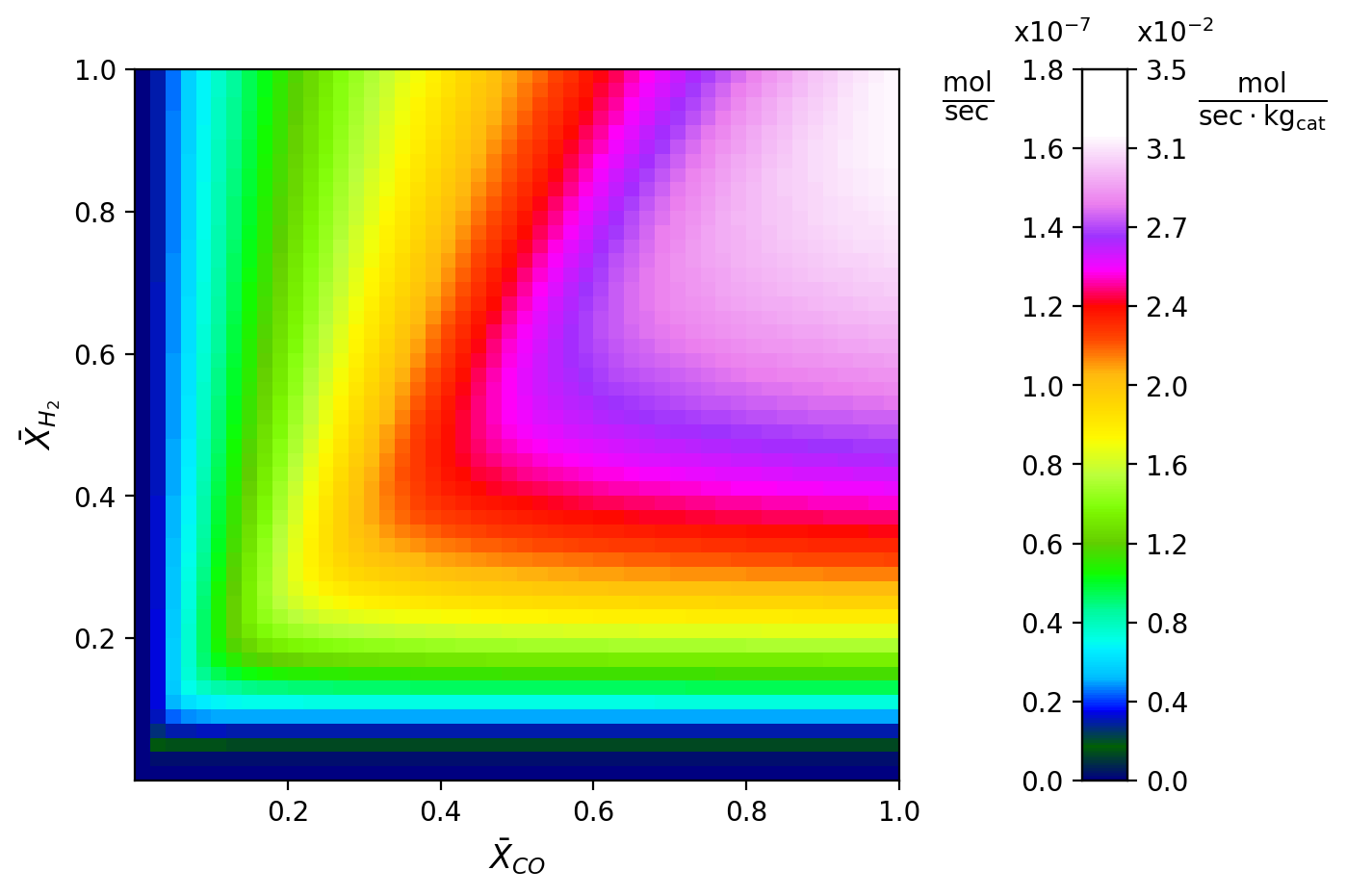]
}
\\
{\textit a}) Total consumption of CO
\\
{
    \halfpagefgraphics[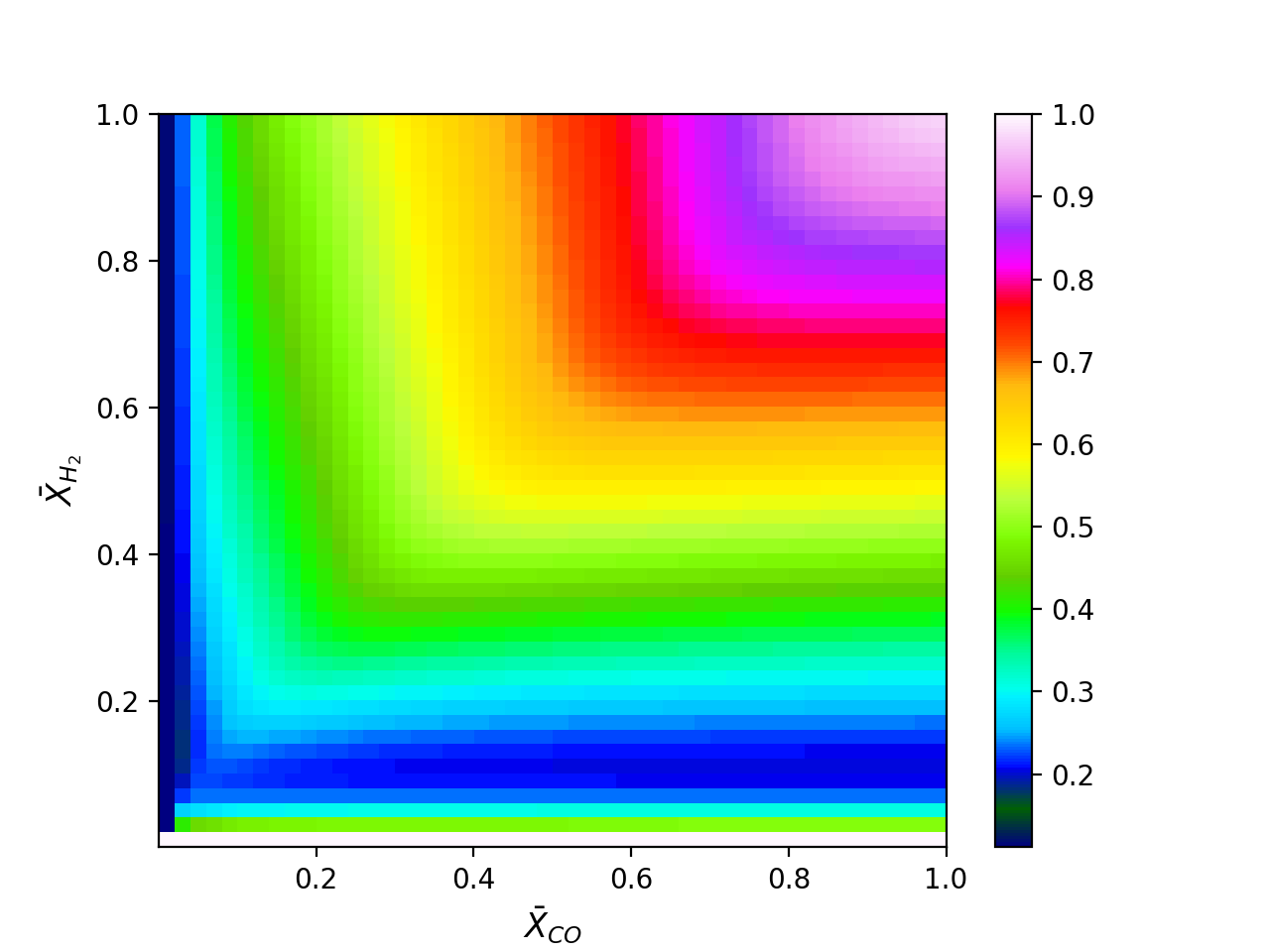]
}
\\
{\textit b}) Pellet effectiveness factor for CO consumption
\\
{
    \halfpagefgraphics[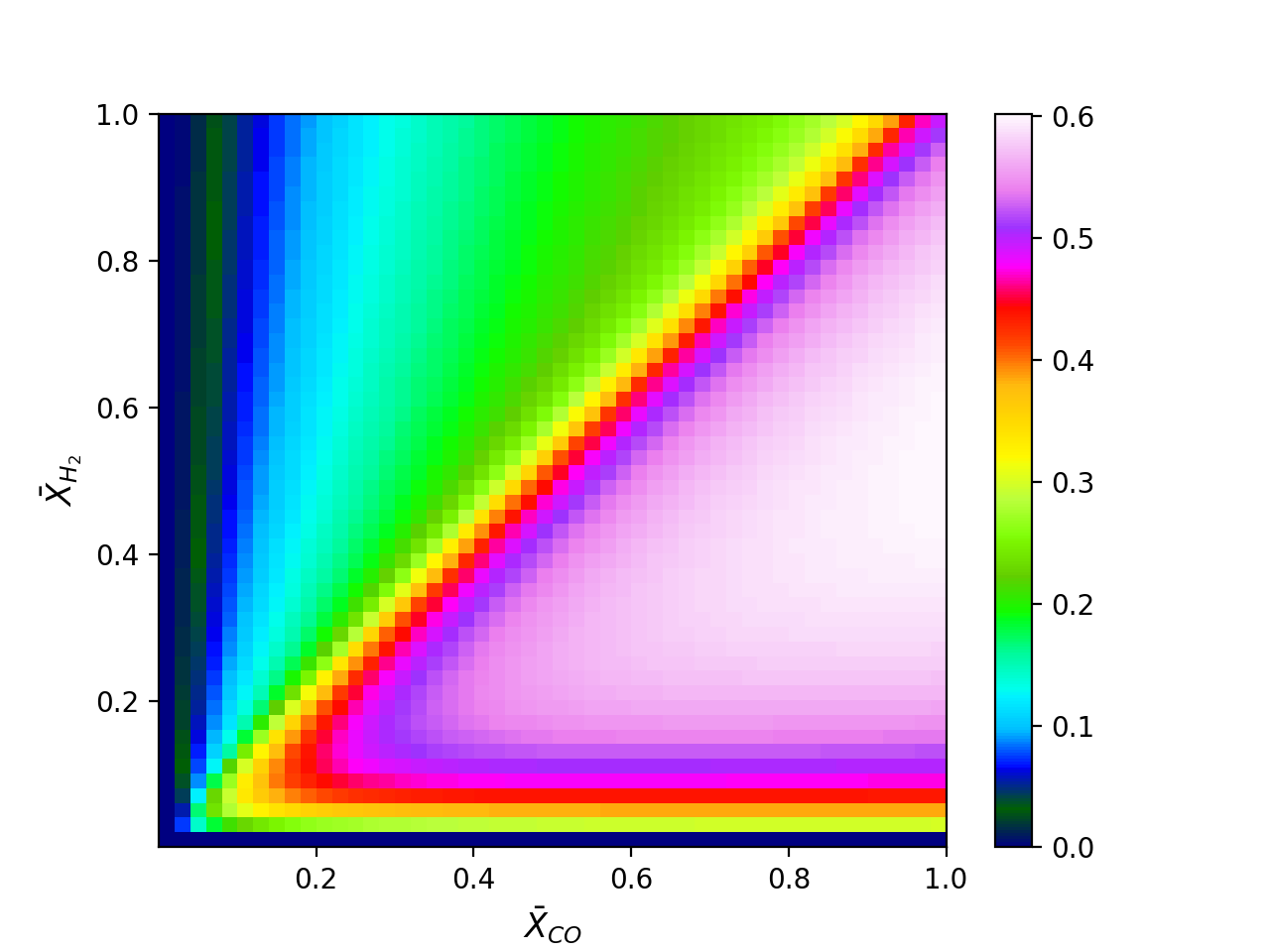]
}
\\
{\textit c}) $\mathrm{C}_{5+}$ fraction
\caption{
  \label{fig:derived-quantities}
  Quantities of chemical interest computed
  as a function of non-dimensionalized reactants pressures
  $  \bar{X}_{CO} =  p_{\mathrm{CO}} / p_{\mathrm{CO}}^{\max}  $ and 
  $\bar{X}_{H_2} = p_{\mathrm{H}_2} / p_{\mathrm{H}_2}^{\max}  $ 
  (cf. (\ref{eq-Xbar-inputs-def})) at 
  $\bar{X}_T = 0.5$, $\bar{X}_{H_2O} = 0.0819 $, 
  based on the solutions of the reaction-diffusion equations obtained with the proposed method accelerated with PINNs: 
  {\textit a}) total consumption of CO (in 
  $\frac{\mathrm{mol}}{\mathrm{sec}}$
and 
$\frac{\mathrm{mol}}{\mathrm{sec} \cdot \mathrm{kg_{cat}}}$
  ), 
  {\textit b}) the pellet effectiveness factor (dimensionless),
  {\textit c}) $\mathrm{C}_{5+}$ fraction (dimensionless)
}
\end{figure}

Obtained dependencies of the considered properties of chemical interest on the partial pressures of CO and $\mathrm{H_2}$ are shown in Fig. \ref{fig:derived-quantities}.

\section{Comparison of initial guess concentration profiles with the final ones}
\label{appendix-profiles-rel-err}

\begin{figure}[h]
\centering{
    \halfpagefgraphics[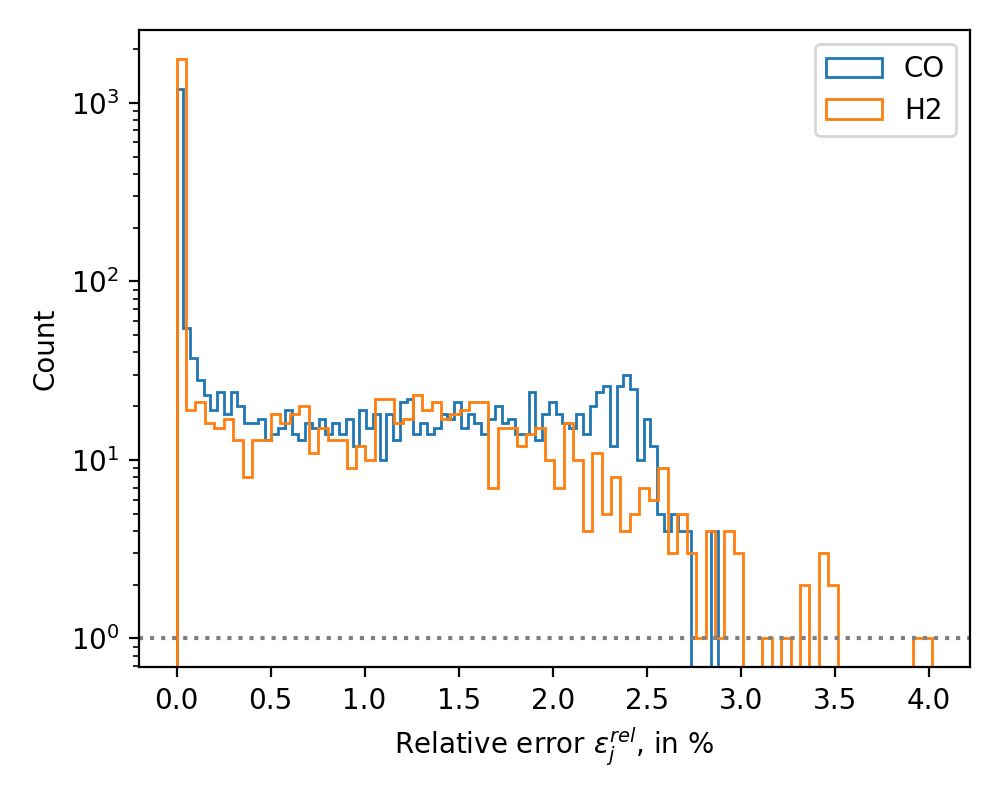]
 }  
  \caption{
  \label{fig:guess-rel-errs}
  The distribution of relative errors 
$
\varepsilon^{\tmop{rel}}_j = \frac{\max_r |w^{final}_j - w^{guess}_j |}{w_j (1)} \cdot 100\%
$ 
($j = \mathrm{CO}, \; \mathrm{H_2}$)
  for the concentration profiles $w_j(r)$ produced by the proposed initial guess generator, as compared to the final solutions
}
\end{figure}

We have compared the initial guesses $w^{guess}_j(r)$ found by the method presented in subsection \ref{appendix-conv-solver-initial-guess} to the ground truth solution $w^{final}_j(r)$, ($j = \mathrm{CO}, \; \mathrm{H_2}$), found by the numerical solver.
These profiles have been found to be rather close, as 
can be illustrated by Fig. \ref{fig:guess-rel-errs} presenting the distribution of the relative errors 
$$
\varepsilon^{\tmop{rel}}_j = \frac{\max_{r \in (0, R_p)} |w^{final}_j - w^{guess}_j |}{w_j (1)} \cdot 100\%
$$
where $w_j (1)$ is the value of the corresponding non-dimensionalized concentration at the pellet boundary,
between $w^{guess}_j(r)$ and $w^{final}_j(r)$, for 2500 instances of boundary conditions.
The mean relative errors for CO and $\mathrm{H_2}$ concentration profiles have been found to be only 0.7\% and 0.4\% respectively.

\RemoveForNow{
Still needs to be done:

{\color{red} Finally, we used the same success rate metrics to explore the effect of both the post-processing transformation we proposed and the problem-specific initialization scheme, in different combinations.}
}

\RemoveForNow{
\begin{table*}[h]
  \begin{tabular}{|l|l|l|}
    \hline
    $N_{\tmop{success}}$ & with post-processing () & original $[S] = 10^y$\\
    \hline
    with initial guess &  & \\
    \hline
    with a constant as initial guess &  & \\
    \hline
  \end{tabular}
  \caption{Number of successfully converged cases}
\end{table*}

}


\section{Peculiarities of the source terms approximations in context of reaction-diffusion equation solver}
\label{appendix-source-terms-approx}

Below we discuss the possible reasons for such a high failure rate with the naive guess and propose an alternative problem-specific approach to creating the initial guess.


For the pellet model, 
the reaction-diffusion 
differential equations which are to be solved can be generalized as
\[ \Delta w = - s (w), \]
i.e., as the Poisson-type equation with the source term $s (w)$ being a non-linear function of the target quantity ‘profile' $w (r)$ (the distribution of either temperature or concentration of one of the substances across the pellet).
For the non-linear equation at hands, numerical methods are always iterative.
Because of that and given the ‘dissipative’ nature of the diffusion and heat transfer processes, in a sense that they always tend to take system to a well-defined steady-state (a non-equilibrium one, in thermodynamic terms), it is quite natural to design the iterations of the solution process such that they mimic the time evolution of the system, starting from some an arbitrary initial guess, compatible with the boundary conditions.
As mentioned in subsection \ref{subsection-eqs-intro}, the problem statement at hands is aimed at finding the stationary concentration profiles, so that timesteps discussed below should rather be treated merely as a numerical heuristics, rather than a physical time.

In context of the iterative solution process, we will use superscripts
$^{\tmop{old}}$ and $^{\tmop{new}}$ to denote the values of the target
quantity at the two subsequent iterations (the moments of ‘time’) and use
column-vector $\mathbf{w}$ to represent the values of the target quantity at
the points of some discrete grid covering the pellet. As the derivatives in
the finite-difference schemes are related to the components of $\mathbf{w}$,
we will use $\mathbf{L}$ to denote the matrix built so that the column-vector
$\mathbf{L} \mathbf{w}$ contains the values of Laplacian of the target
quantity $w$ at the grid points.

Among the finite difference schemes aimed at numerically solving differential
equations (both ODEs and PDEs), it is common to distinguish between the
so-called explicit and implicit schemes \cite{hundsdorfer2013numerical}. Explicit
schemes are easy to implement are computationally less expensive but can
experience convergence issues, constrained by small time-steps pertaining to the stability requirements and therefore require large
number of smaller steps to converge. In contrast to that, stability of
implicit schemes is generally better, they can handle larger steps, but that
is achieved by the need of solving a system of algebraic equations at each
step, thus making entire scheme computationally more demanding and more
difficult to derive from a theoretical perspective.

Accordingly, explicit and implicit schemes for finding
$\mathbf{w}^{\tmop{new}}$ based on $\mathbf{w}^{\tmop{old}}$ can be summarized as
\begin{equation}
  \frac{\mathbf{w}^{\tmop{new}} - \mathbf{w}^{\tmop{old}}}{\tau} = \mathbf{L}
  \mathbf{w}^{\tmop{old}} + \mathbf{s} (\mathbf{w}^{\tmop{old}})
  \label{wnew-wold-explicit-1}
\end{equation}
\begin{equation}
  \frac{\mathbf{w}^{\tmop{new}} - \mathbf{w}^{\tmop{old}}}{\tau} = \mathbf{L}
  \mathbf{w}^{\tmop{new}} + \mathbf{s} (\mathbf{w}^{\tmop{new}})
  \label{wnew-wold-implicit-1}
\end{equation}
respectively. In these equations, $\tau$ is the ‘time’ step (in properly scaled units), and $\mathbf{s} (\mathbf{w})$ is the vector-function returning the values of the source terms $s (w)$ on the grid points. 
If the latter function was linear, using either of (\ref{wnew-wold-explicit-1}) or (\ref{wnew-wold-implicit-1}) would both reduce to solving a system of linear equations at each iteration. 
However, in the pellet modeling problem the
source term function is non-linear and, as a result, using (\ref{wnew-wold-implicit-1}) to find $\mathbf{w}^{\tmop{new}}$ based on $\mathbf{w}^{\tmop{old}}$ requires solving a system of non-linear equations, which complicates the procedure significantly.
Therefore, to make the implicit scheme tractable, it is common to introduce approximations for r.h.s. of (\ref{wnew-wold-implicit-1}) and to express the (small) changes in $\mathbf{s} \left( \mathbf{w}^{\tmop{new}} \right)$ via the changes in $\mathbf{w}$ in a linear manner. 
The coefficients of their relationship can typically be updated at each iteration.
It is the form of these linear approximations that can impact the stability and the performance of the ultimate method significantly.

In the FTS catalyst pellet modeling problem, the source term function \eqref{non-dim-src-term} can be considered as a ‘local’, in a sense that its value at certain point of the pellet depends only on the values of substances concentrations or temperature at the same point.
In other words, $j$-th component of column-vector $\mathbf{s} (\mathbf{w})$ depends only on $j$-th component of column-vector $\mathbf{w}$ (or parameters with respect to which no derivatives are taken in the considered differential equation). 
This feature simplifies the approximation of $\mathbf{s} \left( \mathbf{w}^{\tmop{new}} \right)$, reducing it to a component-wise approximation problem. 
One common option for such approximation could be using the Taylor series expansion for $s_j (w^{\tmop{new}}_j)$ in the vicinity of $w^{\tmop{old}}_j$ and terminating it after the linear terms, viz.
\begin{equation}
  s_j (w_j^{\tmop{new}}) \approx s_j (w_j^{\tmop{old}}) + \left(
  \frac{\partial s_j}{\partial w_j} \right)_{w_j^{\tmop{old}}} \cdummy
  (w_j^{\tmop{new}} - w_j^{\tmop{old}}) \label{eq-swnew-swold-Taylor-linear}
\end{equation}
We have found, however, that this leads to unstable iterations which often fail to converge, especially if the solution contains a region of small concentrations.
Therefore, more delicate approximation scheme is needed.

To this end, we consider the situation when the reactants penetrate into the pellet through its surface from the surrounding medium in a ratio when one of the reactants is consumed much more actively than the others. 
In that case, the closer to the pellet center, the more limited the overall chain of reactions will be by the availability of the actively consumed reactant.
The rate of the elementary reaction in which it is consumed will then become the limiting factor for the the overall rate of chemical transformations. 
It can then be expected that this ‘rate limiting’ elementary reaction will follow first-order kinetics $s = - \alpha \cdummy w$ (which is exactly the case explored in subsection \ref{model-example-subsection}), when the reactant is consumed at the rate that is proportional to the available concentration. 
Such proportionality can be helpful for clarifying the desired properties of approximation for $s_j(w^{\tmop{new}}_j)$.

Under assumption $s = - \alpha \cdummy w$, equation 
$\Delta w = \frac{1}{x^2} \frac{d}{d x} \left( x^2 \frac{d w}{d x} \right) = - s$
can be solved analytically.
The $w (x)$ function satisfying both this equation and the boundary conditions $w' (0) = 0$, $w (1) = 1$, is
$$ w = \frac{1}{\sinh \sqrt{\alpha}} \cdummy \frac{\sinh \left( \sqrt{\alpha}
   \cdummy x \right)}{x} . 
$$
At the pellet center ($x \rightarrow 0$) this concentration is 
$w (0) \rightarrow \frac{\sqrt{\alpha}}{\sinh \sqrt{\alpha}}$, 
so that when $\alpha$ is large enough $w (0) \sim \sqrt{\alpha} \cdummy e^{- \sqrt{\alpha}} \ll 1$ and $w (0)$ becomes vanishingly small, as expected.
In other words, the entire amount of the reactant that penetrates into the catalytic pellet from the outer medium is almost completely consumed in the outermost `shell' of the particle.

Although in a more realistic model $s$ is not linearly proportional to $w$, we can expect that the above described situation can remain qualitatively correct (e.g., by admitting that the ‘effective’ $\alpha$ varying smoothly across the pellet).
Most importantly, the region of small concentrations can still be expected to exist in the realistic setting (e.g., near the pellet center).
Small concentrations, in turn, can possess a difficulty for Taylor series based approximation (\ref{eq-swnew-swold-Taylor-linear}) of $s_j (w^{\tmop{new}}_j)$. 
This can be exemplified by considering $s (w) = A \cdummy w^{0.98}$, which leads to
$\frac{\partial s}{\partial w} = A \cdummy 0.98 \cdummy \frac{1}{w^{0.02}}$,
i.e., the quantity which becomes large when $w \rightarrow 0$.
Consequently, even any small `trial’ changes in concentration which will arise during the iterations of numerical solution method will be ‘amplified’ by large
$\frac{\partial s}{\partial w}$, leading to inaccuracies in $s_j (w^{\tmop{new}}_j)$ approximation. 
For example, $s (w^{\tmop{new}})$ of wrong
sign can appear in the equation, meaning that the reactant is ‘produced’ instead of being consumed. 
Note that getting source terms of a wrong sign as a result of inaccurate $s_j (w^{\tmop{new}}_j)$ approximation is more dangerous than getting incorrect (e.g., too large) source term but with a correct sign (which still corresponds to the consumption of the reactant). 
Indeed, when the chemical transformations are limited by the concentration of one of the reactants, its erroneous `production’ can even make entire system unstable.

With the above considerations, we selected an alternative linear approximation
for $s (w^{\tmop{new}})$, viz.
\begin{multline}
  s (w^{\tmop{new}}) = s (w^{\tmop{old}}) + \frac{s (w^{\tmop{old}}) - s
  (0)}{w^{\tmop{old}} - 0} \cdummy (w^{\tmop{new}} - w^{\tmop{old}}) 
\EqLineBreak
  = \frac{s
  (w^{\tmop{old}})}{w^{\tmop{old}}} \cdummy w^{\tmop{new}},
  \label{sw-linear-approx-we-use}
\end{multline}
where the subscript denoting the number of a grid point has been dropped our for clarity.
This approximation, enabling the use of an implicit numerical
scheme, can be viewed either as devising a ‘local’ value of $\alpha =
\frac{- s}{w}$ at each point of the pellet and then using it to approximate $s (w^{\tmop{new}})$, or as a result of imposing the constraint $s (w = 0) = 0$ and then building a two-point linear approximation for $s$, leading to
$\frac{s (w^{\tmop{old}}) - s (0)}{w^{\tmop{old}} - 0}$
as the value of
$\left(\frac{\partial s}{\partial w} \right)_{w^{\tmop{old}}}$
in (\ref{eq-swnew-swold-Taylor-linear}). 
An essential requirement for applicability of the approximation (\ref{sw-linear-approx-we-use}) is however that $\frac{s (w)}{w}$ ratio remains finite even in small-concentrations regions.
In case NN approximators are involved in $s (w)$ they thus must
ensure correct asymptotic behaviour when $w \rightarrow 0$.

With (\ref{sw-linear-approx-we-use}) used component-wise for $\mathbf{s}
\left( \mathbf{w}^{\tmop{new}} \right)$, the $j$-th component of
(\ref{wnew-wold-implicit-1}) becomes
\begin{equation}
  \left( \frac{s (w^{\tmop{old}}_j)}{w^{\tmop{old}}_j} - \frac{1}{\tau}
  \right) w_j^{\tmop{new}} + (\mathbf{L} \mathbf{w}^{\tmop{new}})_j = -
  \frac{w_j^{\tmop{old}}}{\tau} \label{eq-wnew-from-wold-working-eq}
\end{equation}
which leads to a system of linear equations w.r.t. $\mathbf{w}^{\tmop{new}}$.
Such system is solved with standard numerical routines at each ‘time-step’ iteration of the proposed approach to find $\mathbf{w}^{\tmop{new}}$ from $\mathbf{w}^{\tmop{old}}$.


\end{document}